\documentclass{article}

\usepackage{microtype}
\usepackage{graphicx}
\usepackage{subcaption}
\usepackage{booktabs} 
\usepackage{hyperref}

\usepackage[preprint]{icml2026}

\usepackage{graphicx} 
\usepackage{xcolor} 
\usepackage{amsmath}
\usepackage{amssymb}
\usepackage{mathtools}
\usepackage{amsthm}
\usepackage{booktabs}
\usepackage{amssymb}  
\usepackage{makecell} 
\usepackage{tabularx} 
\usepackage{array} 
\usepackage{minted}

\usepackage{tcolorbox}
\usepackage{enumitem}
\newcommand{\projectname}{\texttt{ECG\_ReasonEval}\ }
\newcommand{\cmark}{\scalebox{1.5}{\checkmark}}
\usepackage[outline]{contour}
\usepackage{bm}
\usepackage{amsmath} 
\usepackage{multirow}

\newcommand{\edit}[1]{{#1}}

\usepackage[capitalize,noabbrev]{cleveref}

\theoremstyle{plain}

\theoremstyle{definition}

\theoremstyle{remark}

\usepackage[textsize=tiny]{todonotes}

\icmltitlerunning{How Well Do Multimodal Models Reason on ECG Signals?}

\begin{document}
\twocolumn[
  \icmltitle{How Well Do Multimodal Models Reason on ECG Signals?}

  \icmlsetsymbol{equal}{*}

  \begin{icmlauthorlist}
    \icmlauthor{Maxwell A. Xu}{uiuc,goog}
    \icmlauthor{Harish Haresamudram}{uiuc}
    \icmlauthor{Catherine W. Liu}{rush}
    \icmlauthor{Patrick Langer}{eth}
    \icmlauthor{Jathurshan Pradeepkumar}{uiuc}
    \icmlauthor{Wanting Mao}{uiuc}
    \icmlauthor{Sunita J. Ferns}{st}
    \icmlauthor{Aradhana Verma}{stan}
    \icmlauthor{Jimeng Sun}{uiuc}
    \icmlauthor{Paul Schmiedmayer}{stan}
    \icmlauthor{Xin Liu}{uw,goog}
    \icmlauthor{Daniel McDuff}{uw,goog}
    \icmlauthor{Emily B. Fox}{stan}
    \icmlauthor{James M. Rehg}{uiuc}
  \end{icmlauthorlist}

  \icmlaffiliation{uiuc}{University of Illinois Urbana Champaign}
  \icmlaffiliation{rush}{Rush University}
  \icmlaffiliation{eth}{ETH Zurich}
  \icmlaffiliation{st}{St. Christopher's Hospital for Children}
  \icmlaffiliation{stan}{Stanford University}
  \icmlaffiliation{uw}{University of Washington}
  \icmlaffiliation{goog}{Google Inc}

  \icmlcorrespondingauthor{Maxwell Xu}{maxu@illinois.edu}

  \icmlkeywords{Machine Learning, ICML}

  \vskip 0.3in
]



\printAffiliationsAndNotice{}  

\begin{abstract}
While multimodal large language models offer a promising solution to the ``black box" nature of health AI by generating interpretable reasoning traces, verifying the validity of these traces remains a critical challenge. 
Existing evaluation methods are either unscalable, relying on manual clinician review, or superficial, utilizing proxy metrics (e.g. QA) that fail to capture the semantic correctness of clinical logic. In this work, we introduce a reproducible framework for evaluating reasoning in ECG signals. We propose decomposing reasoning into two distinct, components: (i) Perception, the accurate identification of patterns within the raw signal, and (ii) Deduction, the logical application of domain knowledge to those patterns. 
To evaluate Perception, we employ an agentic framework that generates code to empirically verify the temporal structures described in the reasoning trace. 
To evaluate Deduction, we measure the alignment of the model's logic against a structured database of established clinical criteria in a retrieval-based approach. This dual-verification method enables the scalable assessment of ``true" reasoning capabilities.
\end{abstract}

\vspace{-4mm}
\section{Introduction}


\begin{table}[]
\centering
\vspace{0mm}
\captionof{table}{
\textbf{Comparison of Reasoning Evaluations.} 
Current methods typically force a trade-off between scalability and depth. Objective proxies (e.g. QA) measure final answer accuracy but ignore the reasoning trace, while Expert Assessment is neither reproducible nor scalable. Furthermore, rigid methods like Step-wise (standard in math) fail in health because they assume a single correct derivation chain. In contrast, clinical diagnosis allows for \textit{multiple valid reasoning paths}: different experts often cite distinct but equally valid sets of features to support the same diagnosis. 
}

\vspace{-1mm}
\label{tab:assessment_comparison_compact}
\setlength{\tabcolsep}{1mm}
\resizebox{\linewidth}{!}{%
\begin{tabular}{l >{\centering\arraybackslash}m{2.0cm} >{\centering\arraybackslash}m{2.0cm} >{\centering\arraybackslash}m{2.0cm} >{\centering\arraybackslash}m{2.3cm} >{\centering\arraybackslash}m{2.0cm}} 
\toprule[1.5pt]

Method & 
\textbf{Applicable to Health Data} & \textbf{Assesses Reasoning Text} & 
\textbf{Inspects Non-Text Modality} & \textbf{Allows Multiple Valid Reasonings} & \textbf{Reproducible} \\
\midrule

\makecell[l]{Step-wise\\Decomposition} & 
& $\cmark$ & $\cmark$ & & $\cmark$ \\
\cmidrule(lr){1-6}

\makecell[l]{Question\\Answering} & 
$\cmark$ & & & & $\cmark$ \\
\cmidrule(lr){1-6}

\makecell[l]{Zero-shot\\Classification} & 
$\cmark$ & & & & $\cmark$ \\
\cmidrule(lr){1-6}

\makecell[l]{n-gram\\Match} & 
$\cmark$ & $\cmark$ & & & $\cmark$ \\
\cmidrule(lr){1-6}

\makecell[l]{Expert\\Assessment} & 
$\cmark$ & $\cmark$ & $\cmark$ & $\cmark$ & \\
\cmidrule(lr){1-6}

\makecell[l]{\projectname\\(Ours)}  & 
$\cmark$ & $\cmark$ & $\cmark$ & $\cmark$ & $\cmark$ \\
\bottomrule[1.5pt]
\end{tabular}%
}
\vspace{-1mm}
\end{table}

\begin{figure*}[!t]
    \centering
    \includegraphics[width=\linewidth]{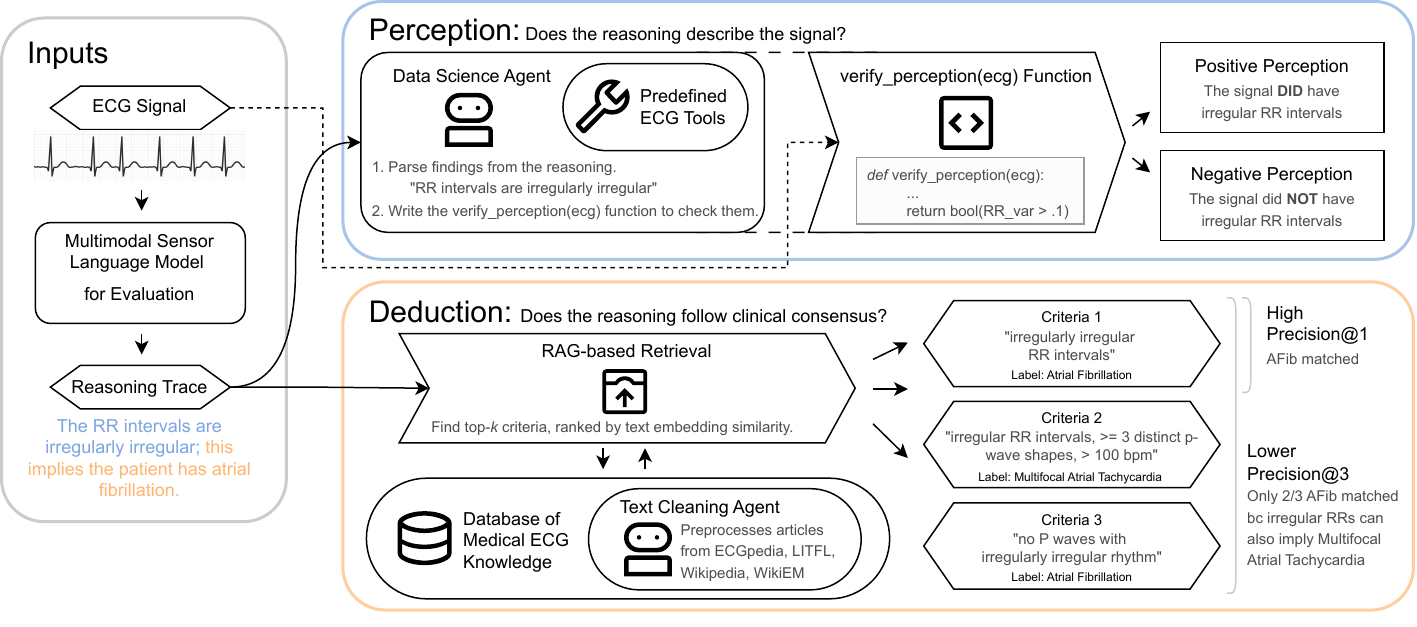}
    \vspace{-6mm}
    \caption{\textbf{\projectname Framework.} We decompose reasoning evaluation into two independent axes: \textit{(i) Perception}, verifying if reasoning is grounded in the signal, and \textit{(ii) Deduction}, evaluating if the logic aligns with clinical consensus. For Perception, we employ a flexible data science agent that dynamically writes and executes Python code to verify specific findings.  For Deduction, we query a medical criteria database with the reasoning trace and check if the retrieved criteria are tagged with the same label as the original signal's ground truth.\looseness -1
}
    \label{fig:overall_framework}
\end{figure*}

Recent advances in Large Language Models (LLMs) have led to their widespread adoption in medical applications, driven by their multimodal data analysis capabilities combined with a natural language interface, which provides a flexible tool for addressing a wide range of health-related questions. 
While the predictive performance of these models has continued to advance in a wide range of clinical domains, key challenges remain. 
A basic question is whether the outputs of an LLM can be trusted for a given clinical application. 
One key tool in addressing end user trust in LLM outputs is the production of \emph{reasoning traces}, sometimes called Chain-of-Thought (CoT) reasoning, which consists of a series of sentences that seem to explain the analysis underlying the model's output. 
While reasoning capabilities have been used extensively 
\citep{wei2022chain, yao2023beyond, chen2024measuring}, there remains significant skepticism whether a language model's reasoning is genuine or merely hallucinated \citep{shojaee2025the}.
Consequently, a fundamental question persists regarding how LLM reasoning should be properly evaluated \citep{lyu2023faithful, lanham2023measuring, wang2023can}.  \looseness -1

While multimodal health models with reasoning capabilities have emerged recently, such as OpenTSLM~\citep{langer2025opentslm} and QoQ-Med~\citep{dai2025qoqmed}, at present there is no systematic way to evaluate the quality of the reasoning output of these models, and as a result it is unclear whether such reasoning is in fact a useful tool for improving the trust and usability of model predictions. 
Benchmarking reasoning capabilities \emph{at scale} is a fundamentally challenging problem, because traces constitute a complex, semantically rich output space and correctness depends on assessing both fidelity to the input data and alignment with standard clinical knowledge. 
Moreover, publicly-available large-scale datasets that combine rich input data with reasoning traces from expert clinicians are rare, making it difficult to obtain reference data for evaluation. 
As a consequence of these challenges, the evaluation of reasoning in both OpenTSLM and QoQ-Med was limited to the annotation of a subset of reasoning traces by expert clinicians. 
This manual approach is neither scalable nor reproducible, and is a significant barrier to fair evaluation of reasoning methods, as it is not practical to re-assemble the expert cohort for every new model. 
To address this gap, researchers often rely on objective proxies for evaluating reasoning, such as question-answering \citep{oh2023ecg, yue2024mmmu} or zero-shot performance \citep{chow2024towards, zhang2025sensorlm}, which only emphasize final-answer accuracy or n-gram match (e.g. BLEU score) \citep{li2024sensorllm}, which does not capture the semantic correctness of reasoning.   \looseness -1

This paper presents the \textit{first scalable approach} to reproducibly evaluating the correctness of multimodal LLM reasoning. 
A key aspect of our approach is to decompose the evaluation task into two separable components: A \emph{Perception} task, which assesses whether the reasoning traces have fidelity with the input data (i.e., does the input possess the properties described by reasoning), and a \emph{Deduction} task which assess the alignment of reasoning with clinical knowledge (i.e., are the assertions made in reasoning correct with respect to gold standard clinical expertise). 
A key strength of this decomposition is that Perception and Deduction can be performed independently, and therefore we don't require a large-scale dataset of biosignals paired with clinician reasoning traces with every combinatorial pairing of perception and deduction. \looseness -1

We develop our novel evaluation approach for the domain of ECG signal analysis to demonstrate a strong initial proof of concept. ECG analysis is an attractive domain for our investigation for two reasons. First, the signal is characterized by standardized morphological features (e.g., P-waves, QRS complexes) explicitly designed for visual inspection by a cardiologist, rendering the Perception task objectively verifiable. Second, the clinical knowledge base for ECG analysis is mature and well-documented, making it feasible to create a well-defined clinical knowledge database for the Deduction task. While ECG reasoning is the focus of this work, we believe our formulation can be extended to a broad set of clinical domains. \looseness -1

Figure~\ref{fig:overall_framework} shows our overall approach, which we illustrate with a concrete example of an ECG reasoning trace: ``The RR intervals are irregularly irregular, which implies Atrial Fibrillation''. 
To validate this assertion, we must first verify \textit{(i)} Perception: Are the R-peaks in the actual signal irregularly spaced? Then, we must verify \textit{(ii)} Deduction: Do irregular RR intervals necessarily imply Atrial Fibrillation? 
For Perception, we develop an agentic framework in which an LLM equipped with specialized tools writes code that empirically verifies that the temporal structures described are in fact present in the signal. 
Next, we evaluate Deduction by comparing the reasoning assertions to the established medical literature. 
Leveraging the fact that medical diagnoses are 
exhaustively documented, we first parse online cardiology resources into a structured reference database. 
Using a Retrieval approach, we can then quantify the validity of a reasoning trace by measuring how often it can retrieve criteria are tagged with the same label as its ground truth label. \looseness -1

Our key contributions are:

\vspace{-3mm}
\begin{enumerate}[leftmargin=*,  noitemsep] 

\item {A Reasoning Evaluation Framework:} We introduce \projectname\footnote{Code and Data available here: url omitted for public arxiv.}, 
the first reproducible framework for evaluating the semantic correctness of reasoning traces in multimodal time-series models. Our automated approach provides a reliable standard that obviates the need for unscalable manual review for every new model.\looseness -1

\item {Novel Decomposition Methodology:} We decompose reasoning into \textit{Perception} (grounding in signal features) and \textit{Deduction} (alignment with clinical consensus). We operationalize these independently using agentic verification for signal fidelity and retrieval for clinical logic. Our experiments confirm the framework's robustness, demonstrating its ability to audit models and even identify errors in cardiologist annotations. \looseness -1

\item {Insights on Multimodal Models:} TSLMs show good Perception but lack medical knowledge for Deduction. Frontier LLMs do well in Deduction but seem to hallucinate signal features post-hoc. This confirms that high predictive accuracy does not imply trustworthy reasoning. \looseness -1
\end{enumerate}

\section{Related Work} \label{appendix:relwork}

\paragraph{Time-Series Language Models.}
Integrating time-series with LLMs remains challenging. Common approaches include tokenizing signals for forecasting \cite{gruver2023large, ji2024hargpt} or projecting embeddings for classification \cite{pillai2025time2lang, ye2024medualtime, li2024sensorllm}, though these often lack text generation capabilities. 
Recent work has shifted towards unified architectures capable of complex reasoning and question-answering \cite{wang2025chattime, xie2024chatts}. 
QoQ-Med is a model \cite{dai2025qoqmed}, which integrates many heterogeneous modalities, such as X-Ray, CT Scan, MRI, and ECG, and generates reasoning text for predicting labels with Relative Policy Optimization. 
Models such as GEM \cite{lan2025gem} and OpenTSLM \citep{langer2025opentslm} have been trained on ECG signals, and both use  external models (e.g. GPT4o) to generate viable reasoning traces for training. \looseness -1

\paragraph{Evaluating Reasoning.}
General reasoning benchmarks typically rely on final-answer accuracy in QA formats (e.g., Humanity's Last Exam \cite{phan2025humanity}) or step-wise decomposition for math problems \cite{zhou2025mpbench}. 
However, unlike the rigid logic of mathematics, clinical interpretation is often ambiguous and will not have consistency on annotated intermediate steps required for such granular evaluation \citep{davies2019healthcare}. 
Physicians often use different lines of reasoning to reach a diagnosis \citep{norman2017causes} with high inter-rater variability \citep{magee2014reliability, cook2020accuracy, magee2014reliability}.


\section{Methodology}
\label{sec:methodology}

\begin{figure}[!t]
    \vspace{-3mm}
    \centering
    \includegraphics[width=\linewidth]{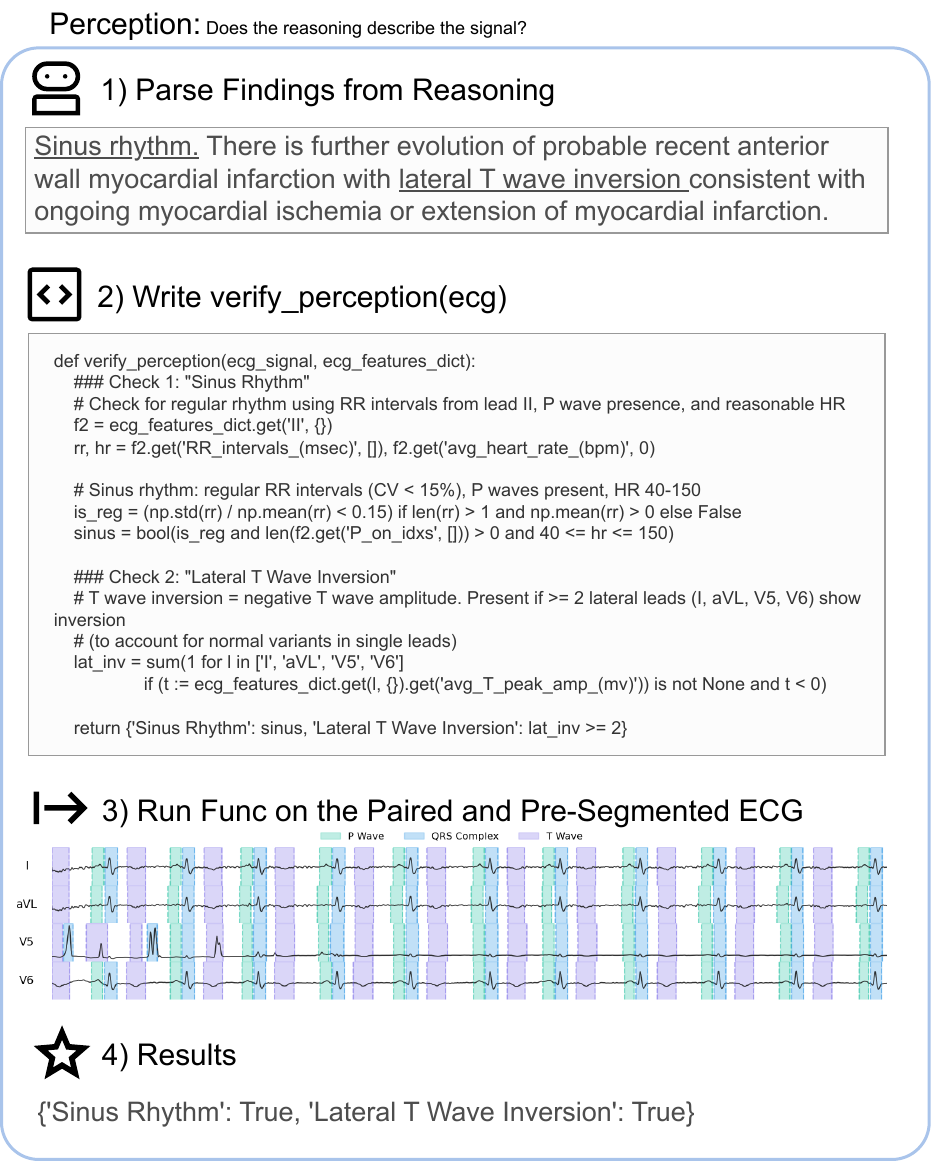}
    \vspace{-6mm}
    \caption{\textbf{The Perception Pipeline.} First, our data science agent extracts discrete, verifiable findings from the reasoning trace. It then generates executable Python code to empirically verify these claims against the raw ECG signal. This process is augmented by a segmentation tool, which lowers the code generation complexity by providing pre-computed wave delineations. The pipeline outputs a boolean verification status for each finding, determining if the reasoning describes the signal accurately.}
    \label{fig:perceptionexample}
    \vspace{-6mm}
\end{figure}

Here, we describe the Perception and Deduction phases of our framework, which is illustrated in \Cref{fig:overall_framework}. Please refer to Appendices \ref{appendix:perception} and \ref{appendix:deduction} for additional details (e.g., prompts). \looseness -1


\subsection{Perception Phase} \label{sec:perception}
This phase evaluates if the reasoning describes the actual signal. 
The inference is visualized in \Cref{fig:perceptionexample}. \looseness -1

\textbf{Motivation.} 
The goal 
is to autonomously determine if a reasoning trace refers to observable patterns empirically present in the signal. 
The feasibility of using LLMs to generate code for data analysis is well-established \citep{gu2024blade, hu2024infiagent, merrill2024transforming, wu2024daco, yin2023natural, zhang2025datascibench, zhang2024benchmarking}. In the health domain specifically, \citet{heydari2025anatomy} successfully applied this approach to answer complex, open-ended queries  (e.g. "Has my average daily step count significantly changed in the past year?") on a user's wearable data for a personal health coach \citep{Google2025personalhealthcoachpha}. By leveraging these capabilities, our agent can programmatically verify the existence of specific ECG features detailed in a given reasoning trace. 

That being said, applying this approach is not straightforward due to the deep domain expertise involved in designing data science agents. 
An innocuous statement, such as ``the RR interval are irregularly irregular", while straightforward to a physician, may not be obvious to those without a background in ECGs. 
It would require the data science agent to recognize that RR intervals represent the time between R peaks and that R peaks are the highest points of an ECG signal, representing the rapid electrical activation (depolarization) of the heart's ventricles, in addition to understanding the normal thresholds for regularity. 

\begin{table}[t]
    \vspace{-2mm}
\centering
\setlength{\tabcolsep}{2mm}
\caption{
\textbf{Assessment of \projectname Perception.} We evaluate our Perception pipeline's reliability using two protocols: \textit{Supporting Assessment}, which measures the verification rate of ground-truth cardiologist notes, and \textit{Adversarial Assessment}, which tests robustness against hallucination by inverting clinical descriptors (e.g. Wide $\leftrightarrow$ Narrow QRS). The strong performance gap between these settings confirms that \projectname effectively distinguishes between present and absent signal features. 
    }
        \vspace{-2mm}
\resizebox{\columnwidth}{!}{%
\begin{tabular}{
    l
    >{\centering\arraybackslash}m{1.8cm}
    >{\centering\arraybackslash}m{1.8cm}
    >{\centering\arraybackslash}m{1.8cm}
    >{\centering\arraybackslash}m{1.8cm}
}
\toprule
\textbf{Perception Assessment} &
\multicolumn{3}{c}{\textbf{\contour{black}{$^{\uparrow}$}Supporting Assessment}} &
\textbf{\contour{black}{$^{\downarrow}$}Adversarial Assessment} \\
\cmidrule(lr){2-4} \cmidrule(lr){5-5}
& \textbf{Global Acc}
& \textbf{Acc@ Thresh50\%}
& \textbf{Acc@ Thresh100\%}
& \textbf{Acc@ Thresh100\%} \\
\midrule
\textbf{\projectname}
& $0.830{\scriptscriptstyle \pm .000}$
& $0.939{\scriptscriptstyle \pm .002}$
& $0.669{\scriptscriptstyle \pm .009}$
& $0.124{\scriptscriptstyle \pm .005}$ \\
\midrule
\multicolumn{5}{l}{\textit{Ablations}} \\
w/o Parsing Findings
& $0.631{\scriptscriptstyle \pm .014}$
& --
& $0.631{\scriptscriptstyle \pm .014}$
& $0.138{\scriptscriptstyle \pm .003}$ \\
w/o Unified Agent
& $0.715{\scriptscriptstyle \pm .006}$
& $0.842{\scriptscriptstyle \pm .005}$
& $0.463{\scriptscriptstyle \pm .017}$
& $0.088{\scriptscriptstyle \pm .008}$ \\
w/o DL Segment Tool
& $0.754{\scriptscriptstyle \pm .010}$
& $0.913{\scriptscriptstyle \pm .006}$
& $0.581{\scriptscriptstyle \pm .003}$
& $0.128{\scriptscriptstyle \pm .005}$ \\
w/o Tools
& $0.688{\scriptscriptstyle \pm .020}$
& $0.816{\scriptscriptstyle \pm .017}$
& $0.463{\scriptscriptstyle \pm .019}$
& $0.141{\scriptscriptstyle \pm .003}$ \\
w/o Strong LLM
& $0.506{\scriptscriptstyle \pm .000}$
& $0.635{\scriptscriptstyle \pm .007}$
& $0.244{\scriptscriptstyle \pm .011}$
& $0.027{\scriptscriptstyle \pm .010}$ \\
\bottomrule
\end{tabular}
}     \vspace{-6mm}
\label{tab:perception_valid}
\end{table}

\textbf{Architectural Design.}

\emph{A) Multi-task Single-Agent.} We implement a single-agent system that executes two distinct tasks within a single prompt context: 
\textit{(i) }extraction and \textit{(ii) } verification. 
First, the agent parses the unstructured reasoning trace into discrete, verifiable findings (e.g. [``inconsistent RR intervals", ``ST elevation $>$ 0.1mV in V1"]). 
As shown in Table \ref{tab:perception_valid}, extraction significantly improves performance by forcing the model to explicitly plan its verification logic before generating code, while simultaneously offering granular interpretability.

Following this, the agent utilizes a toolset described below 
to write and execute Python verification functions. 
We incorporate a self-correcting feedback loop where execution errors prompt the agent to debug its own code. 
Persistent failure in generating executable logic is treated as a negative result, implying the original reasoning trace was too abstract to be empirically verified. Notably, this failure mode never occurred with \projectname across all evaluations: it was only observed within the ablations. 


Our ablation experiments (Table \ref{tab:perception_valid}) confirm that this single-agent approach outperforms complex multi-agent baselines for both extraction and verification, which are impacted by coordination friction and `tool coordination tradeoff' \citep{kim2025towards}, leading to reduced performance.

\emph{B) Specialized Tooling.} 
We equip the agent with domain-specific tools to lower the level of code generation complexity. 
As we rely on the agent's generated code as a ground-truth evaluation mechanism, minimizing logic errors in the verification process is critical. 

ECG interpretation relies heavily on segmenting distinct portions of the signal. 
For example, common findings in cardiologist notes such as ``prolonged PR interval", ``wide QRS complex", and ``ST segment elevation" all depend on the accurate delineation of the P, Q, R, S, and T ECG wave boundaries. 
However, reliably performing segmentation is non-trivial. 
Classical rule-based algorithms \citep{emrich2024physiology, alarcon2025optimization} often fail on pathological data despite high benchmark scores on healthy signals \citep{joung2024deep}. 
As shown in Table \ref{tab:perception_valid}, these methods degrade performance by frequently hallucinating features. 
We instead equip our agent with a segmentation tool based on a SOTA DL model explicitly trained on diverse clinical cohorts, resulting in robustness across pathologies \citep{park2025semisegecg}. 
This idea can be easily generalized to non-ECG-specific settings. 
For example, given a radiology setting, SOTA vision models can be provided to identify abnormal masses \looseness -1.


\begin{figure}
    \vspace{-2.5mm}
    \centering
    \includegraphics[width=0.8\linewidth]{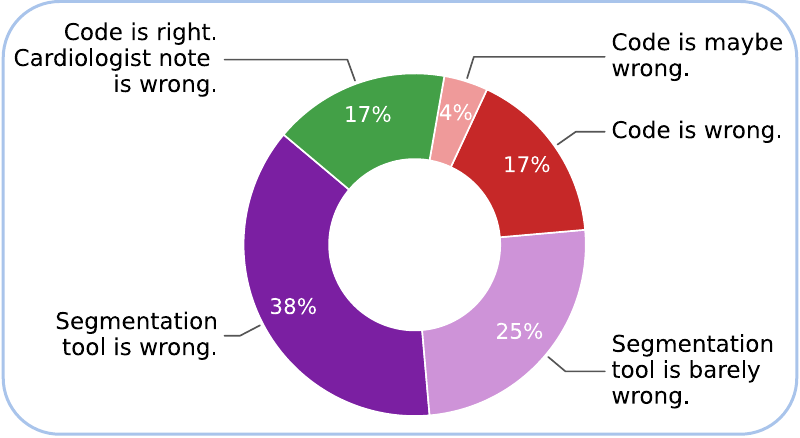} \\
    \fontsize{7.5}{8.5}\selectfont 
    \textit{``Code is right; Cardiologist note is wrong"}: Instances where the ground truth note incorrectly described the ECG. The note is marked wrong after unanimous consensus across our physician team. \\
    \textit{``Code is maybe wrong"}: Ambiguous cases where missing demographic data (e.g. age) prevented definitive verification of "normal" limits. \\
    \textit{``Tool is barely wrong"}: Borderline segmentation discrepancies where values marginally exceeded strict thresholds (e.g. QRS of 121ms vs. 120ms).
    \vspace{-2mm}
    \caption{
    \textbf{Causes of Perception Failure.} With our team of physicians, we manually examined the failed reasoning traces in the val set to categorize error sources. 
    The results suggest Perception is highly reliable and capable of auditing errors in human annotations. \looseness -1
    }
    \vspace{-6mm}
    \label{fig:perception_failure}
\end{figure}

\emph{C) Data Science Agent's LLM Backbone.} 
We utilize Claude 4.5 Opus \citep{anthropic2024claudeopus45}, the current state-of-the-art coding model. Crucially, Anthropic does not use API interactions to train models or for routine human reviews of data. 
Thus, they are compliant with MIMIC-IV's Credentialed Data Use Agreement \citep{mitlcp2024mimicivailicense}, which allows us to deploy the data science agent on MIMIC-IV-ECG \citep{gow2023mimicivecg}. 
We also evaluate Claude 3 Haiku in Table \ref{tab:perception_valid} (row ``w/o Strong LLM"), and find that simpler models lack the necessary capabilities to function effectively as the agent backbone \looseness -1.



\textbf{Metrics.} 
Our approach returns a boolean True/False for each finding within a reasoning trace, which enables evaluation on a per-reasoning basis. 
We utilize two metrics: \textit{(i)} Accuracy@Threshold$P$\%, which is the percentage of reasoning traces with at least $P$\% of the findings within verified. 
In other words, when $P=100$, then a reasoning is only marked as ``correct" when 100\% of the findings are verified; and \textit{(ii)} Global Accuracy, which is 
the total percentages of verified findings across all reasoning traces.


\textbf{Assessment.} 
To ensure our data science agent is able to analyze a wide variety of clinical interpretations found in real-world settings, we validate it on MIMIC-IV-ECG \citep{gow2023mimicivecg}. 
It contains cardiologist reports with paired 12-lead ECG recordings (10s at 500 Hz). 
To reduce excessive computational costs, we 
validate on a random subset of 1,000 Note+ECG pairs from 981 unique patients. 
We employ a random 10/90 val/test split with val being used for tuning the prompt and other parameters, and report test performance. 
We perform two types of assessments: 

\emph{i) Supporting Assessment:} 
Here, the agent is tasked with programatically verifying that the morphological and rhythmic features described in these expert notes are grounded in the raw ECG signals. 
High Perception rate indicates that our agent possesses sufficient signal processing logic to verify and corroborate cardiologists' observations, and consequently, to accurately ground clinical concepts in time-series data. 
Ideally, Acc@Thresh100\%=1.0 because we trust that the cardiologist notes is describing the ECG signal correctly. \looseness -1

\vspace{-2mm}
\emph{ii) Adversarial Assessment:} 
To ensure the agent does not blindly validate all inputs, we explicitly flip distinct clinical descriptors found in the notes to their opposite meanings.
For example, we change ST elevation to ST depression (see Table \ref{tab:app:adv_map} in the Appendix for the full list).
As these altered statements contradict the ground truth, we expect the agent to return negative Perception results. 
This serves as a negative control, ensuring the agent can correctly identify when a textual claim does not exist in the waveform.

\emph{Results:} Table \ref{tab:perception_valid} shows \projectname accurately identifies present signal features with high accuracies during the supporting assessment and low ones during adversarial. 
Our team of physicians performed a manual review of the discrepancies when our agent deviates on the validation set, shown in \Cref{fig:perception_failure}. 
The high incidence of segmentation tool errors (63\%) is encouraging, as it implies that the core Perception methodology is sound. Then, because our framework is modular, performance can be directly improved by adopting state-of-the-art models as they emerge. The ``Code is maybe wrong" cases highlight a clear direction for future work, specifically the need to integrate additional clinical context, an essential feature for benchmarking against models that rely on patient metadata. Instances where the ``Cardiologist note is wrong" are analyzed in Section \ref{sec:discussion}.

\subsection{Deduction Phase} \label{sec:deduction}

\begin{figure}[!t]
    \vspace{-3mm}
    \centering
    \includegraphics[width=\linewidth]{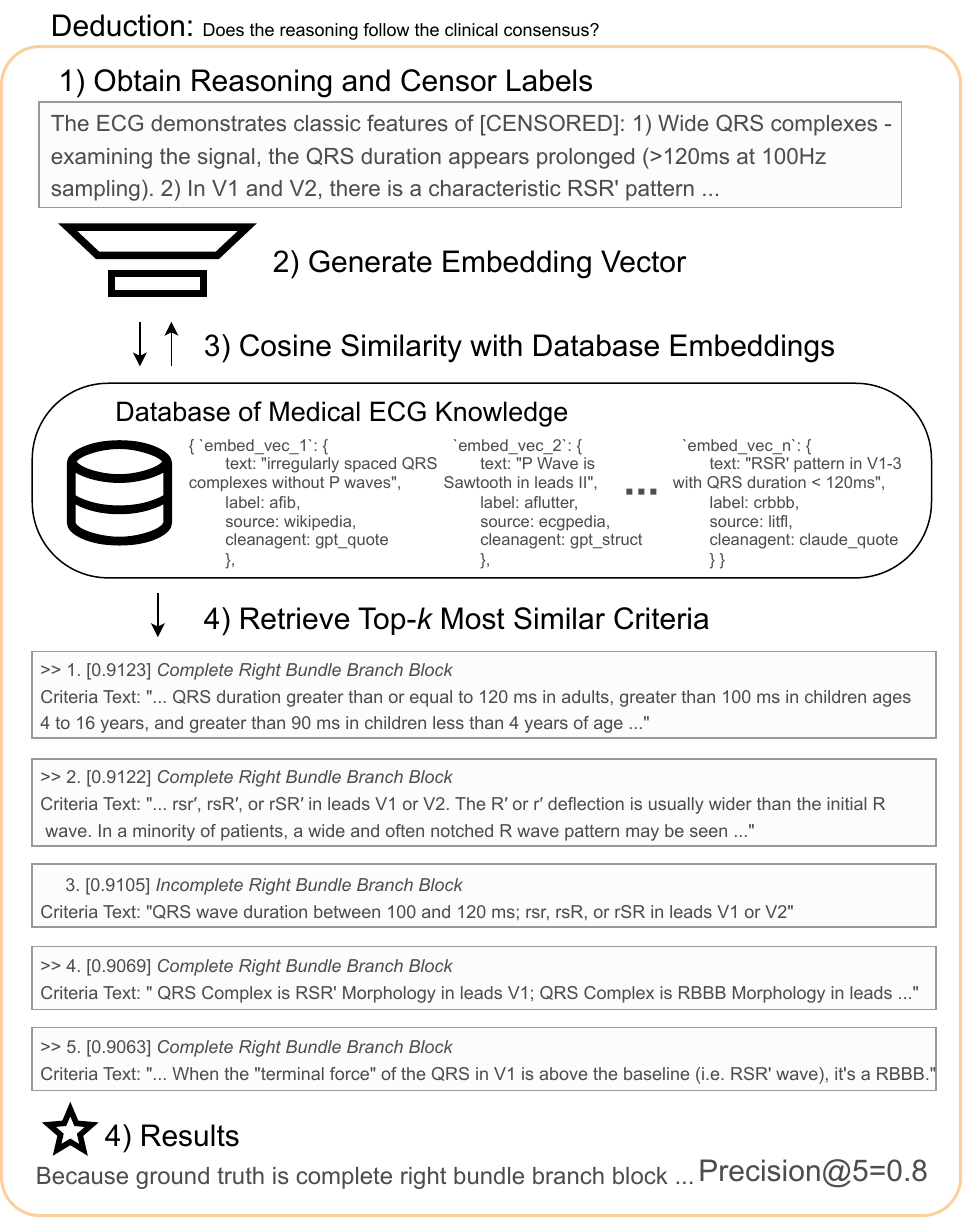}
    \vspace{-6mm}
    \caption{\textbf{The Deduction Pipeline.} The input reasoning trace is first censored of its final diagnostic label, embedded using the Gemini model, and used to query our database of diagnostic criteria. We retrieve the top-$k$ most semantically similar articles and calculate Precision@$k$ against the ground truth. A high score indicates that the reasoning trace accurately maps to the correct pathology according to established medical standards, mirroring how physicians cross-reference diagnostic criteria.}
    \label{fig:deduction_pipeline}
        \vspace{-4mm}
\end{figure}

First, a database must be constructed containing a set of criteria for every diagnostic label. 
This is accomplished by scraping raw text from authoritative cardiology resources (e.g. \href{https://litfl.com/ecg-library/}{\color{black}LITFL}), and then passing it to the Text Cleaning agent. 
The Deduction pipeline is visualized in \Cref{fig:deduction_pipeline}. 

\textbf{Architectural Design} 

\emph{A) Label Compilation.} 
We first establish a comprehensive clinical taxonomy to index the diagnostic criteria in our knowledge base. 
We aggregate the complete set of diagnostic labels from our downstream evaluation datasets, ECG-QA and MIMIC-IV-ECG. 
This results in a diverse vocabulary of clinically validated terms, spanning high-level diagnostic conditions (e.g. {Infarction}), rhythm disturbances (e.g. {AFib}), and form abnormalities (e.g. {inverted T wave}).

\emph{B) Raw Text Compilation.} 
To populate our knowledge base, we utilize the Google Search API to retrieve the top-5 most relevant articles for each label (e.g. query: ``atrial fibrillation site:litfl.com'') from the following domain-specific sources: \textit{(i)} LITFL ECG Library; \textit{(ii)} Wikipedia; \textit{(iii)} ECGPedia; and \textit{(iv)} WikiEM. 
We detail these sources and their license in \Cref{sec:appendix:knowledge_sources} of the Appendix.
We employ site-specific parsing scripts to convert the raw HTML content into standardized Markdown, creating a structured corpus of reference text.
This strategy leverages the rich availability of high-quality, open-access medical curricula that serve as the de facto training standard and reference materials for modern physicians. 
As shown in Table \ref{tab:physician_validation}, removing any individual source  degrades performance across physician evaluations. \looseness -1





\emph{C) Text Cleaning Agent.} 
It parses the scraped articles and isolates specific sets of diagnostic criteria for a given condition. 
This step is critical because raw search results often contain significant noise, navigational text, or tangentially relevant information. 
As demonstrated in Table \ref{tab:physician_validation} (row ``w/o Cleaning Agent''), performance degrades significantly across all physician baselines when this filtration step is removed, highlighting its necessity.


To capture the full nuance of clinical consensus, 
we deploy four distinct variations of the Text Cleaning Agent, defined by the combination of their extraction strategy and LLM backbone. 
We utilize two complementary strategies: \textit{(i)} Exact Quote Extraction, which identifies verbatim excerpts that define the diagnosis; and \textit{(ii)} Structured Synthesis, which summarizes and standardizes the criteria into a consistent template. By instantiating each strategy with two state-of-the-art models, Claude 4.5 Opus \citep{anthropic2024claudeopus45} and GPT 5.2 Pro \citep{openai2025gpt52}, we generate four diverse knowledge representations per article (2 strategies × 2 models).


Table \ref{tab:physician_validation} indicates that ensembling the outputs yields the highest performance. 
By combining exact quotes with standardized summaries across different LLM backbones, we create a richer, more robust semantic target for retrieval. This ensemble approach outperforms any single extraction method or model,
confirming that diversity in knowledge representation is key to aligning with expert consensus. 

\emph{D) Retrieval of Top-$k$ Criteria.} 
To prevent data leakage, we first censor the input reasoning trace by removing all explicit mentions of the final diagnostic prediction. 
Subsequently, we extract embeddings of this sanitized trace using Gemini's text embedding model \citep{team2023gemini}, and compute its cosine similarity against the pre-computed vector index of our knowledge base. 
We retrieve the top-$k$ most relevant criteria, and verify alignment by checking if their label matches the ground-truth label of the input data.

\begin{table}[t]
\vspace{-2mm}
\centering
\caption{\textbf{Assessment of \projectname Deduction.} 
We evaluate the alignment of our retrieved medical references used during Deduction against diagnostic criteria independently authored by three physicians. The full \projectname pipeline consistently performs well with high Precision@$k$, confirming that our approach effectively captures  nuanced ``clinical logic".}
\vspace{-2mm}
\label{tab:physician_validation}
\resizebox{\linewidth}{!}{%
\setlength{\tabcolsep}{1pt}
\begin{tabular}{lccccccccc}
\toprule
\makecell[l]{\textbf{Deduction} \\ \textbf{Assessment}} &  
\multicolumn{3}{c}{\makecell{\textbf{Physician A}}} &  
\multicolumn{3}{c}{\makecell{\textbf{Physician B}}} &  
\multicolumn{3}{c}{\makecell{\textbf{Physician C}}} \\
\cmidrule(lr){2-4} \cmidrule(lr){5-7} \cmidrule(lr){8-10}
& {\contour{black}{$^{\uparrow}$}Precis} & {\contour{black}{$^{\uparrow}$}Precis} & {\contour{black}{$^{\uparrow}$}Precis} 
& {\contour{black}{$^{\uparrow}$}Precis} & {\contour{black}{$^{\uparrow}$}Precis} & {\contour{black}{$^{\uparrow}$}Precis} 
& {\contour{black}{$^{\uparrow}$}Precis} & {\contour{black}{$^{\uparrow}$}Precis} & {\contour{black}{$^{\uparrow}$}Precis} \\
& {@1} & {@5} & {@10}
& {@1} & {@5} & {@10}
& {@1} & {@5} & {@10} \\
\cmidrule(lr){1-10}

\textbf{\projectname} & 0.83 & 0.76 & 0.69 & 0.79 & 0.78 & 0.71 & 0.79 & 0.75 & 0.70 \\
\midrule

\multicolumn{10}{l}{\textit{Ablations}} \\
w/o Quote Extract     & 0.73 & 0.66 & 0.63 & 0.83 & 0.70 & 0.65 & 0.72 & 0.70 & 0.64 \\
w/o Struct. Extract   & 0.77 & 0.65 & 0.59 & 0.74 & 0.73 & 0.65 & 0.77 & 0.73 & 0.66 \\
w/o two LLMs          & 0.76 & 0.72 & 0.65 & 0.72 & 0.70 & 0.66 & 0.74 & 0.69 & 0.65 \\
w/o Cleaning Agent    & 0.40 & 0.42 & 0.35 & 0.37 & 0.40 & 0.35 & 0.47 & 0.42 & 0.36 \\
\midrule

\multicolumn{10}{l}{\textit{Sources}} \\
w/o LITFL             & 0.77 & 0.70 & 0.64 & 0.80 & 0.68 & 0.65 & 0.70 & 0.68 & 0.65 \\
w/o ECGPedia          & 0.83 & 0.77 & 0.71 & 0.78 & 0.77 & 0.69 & 0.79 & 0.73 & 0.69 \\
w/o Wikipedia         & 0.76 & 0.68 & 0.63 & 0.71 & 0.76 & 0.68 & 0.74 & 0.71 & 0.68 \\
w/o WikiEM            & 0.78 & 0.70 & 0.65 & 0.76 & 0.80 & 0.70 & 0.77 & 0.73 & 0.69 \\
\bottomrule
\end{tabular}
}
\vspace{-6mm}
\end{table}

\textbf{Metrics.} 
We employ Precision@$k$, which measures the proportion of the top-$k$ retrieved articles that match the correct ground-truth label. This effectively grades the {specificity} of the reasoning. 
A higher precision implies the model's logic is consistently mapping to the correct disease rather than ambiguous or unrelated conditions. \looseness -1

\textbf{Assessment.} 
To validate Deduction as a reliable proxy for clinical consensus, our physician team were tasked to independently define diagnostic criteria for each label within our database. 
Then, each of their defined criteria was run on our Deduction pipeline to assess whether our pipeline is able to correctly align the database criteria to the physician-defined criteria. \looseness -1

The team is composed of three practicing physicians with varying levels of expertise: an internal medicine resident (A), a cardiology fellow (B), and a Division Chief of Cardiology (C). We use a 1:2 val/test split. Specifically, we tune our text cleaning agent using the criteria provided by Physician A and subsequently test the agent's performance against the criteria authored by the Physicians B and C.

\emph{Results.} Table \ref{tab:physician_validation} demonstrates strong alignment between our database and expert consensus, achieving $\sim$0.8 Precision@1 across all raters. Further discussion in Section \ref{sec:discussion}.

\section{Experimental Evaluation}
\begin{figure*}
    \vspace{-3mm}
    \centering
    \includegraphics[width=\linewidth]{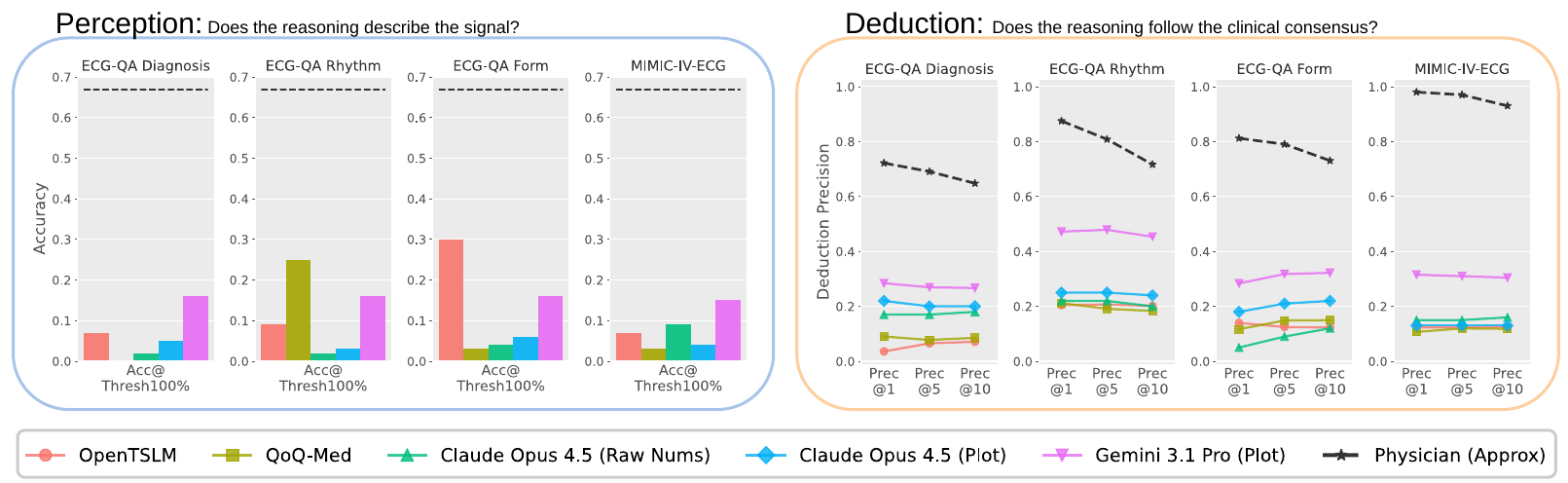}
    \vspace{-6mm}
    \caption{\textbf{Summary of Baselines with \projectname on Perception and Deduction.} We see all models \edit{perform} much worse than the physician performance across both perception and deduction. \edit{While} the TSLMs (OpenTSLM and QoQMed) that were trained with explicit time-series adapters \edit{often} do better in perception to capture time-series features correctly, \edit{frontier VLMs like} Claude Opus 4.5 Plot \edit{and Gemini 3.1 Plot are} able to harness \edit{their} world knowledge to align with the broader clinical literature, achieving stronger deduction performance. \edit{Most notably, Gemini establishes the strongest overall baseline across both perception and deduction metrics.}}
    \vspace{-3mm}
    \label{fig:overall_performance}
\end{figure*}

In \Cref{sec:methodology}, we developed and validated our framework for evaluating the ability of multimodal language models to reason on ECG signals. Here, we describe our experimental setup and the reasoning models we evaluate on. 

\textbf{Multimodal Reasoning Models.} We compare both open and close sourced models, spanning both Time-Series Language Models (TSLMs) as well as two flagship LLMs, Anthropic's Claude Opus 4.5 \citep{anthropic2024claudeopus45} \edit{and Google's Gemini 3.1 Pro \citep{google2025gemini3}}. All models are evaluated in a zero-shot fashion.
For the TSLMs, we utilize QoQ-Med \cite{dai2025qoqmed} and OpenTSLM \cite{langer2025opentslm}. 
QoQ-Med is a multi-modal clinical foundation model capable of utilizing any combination of medical imaging (e.g. x-rays, fundus images, pathology), EHR data, text, and ECG, for medical tasks such predicting diagnoses.
On the other hand, OpenTSLM is focused on reasoning over medical text and time-series data only. 
Both models are also open source and we utilize their original prompts.
For Claude Opus 4.5, we provided ECG signals in two formats: raw numerical data and plotted ECG waveforms, allowing us to assess how frontier models can be applied to ECG analysis. \edit{For Gemini 3.1, we focus on only the plot variation, as that demonstrated stronger performance.} \looseness -1

\textbf{Tasks.} These models are evaluated on four tasks (full details in \Cref{appendix:taskdetails}): \textit{(i)} predicting ICD-10 diagnostic codes on MIMIC-IV; \textit{(ii)} predicting diagnostic, \textit{(iii)} rhythm, and \textit{(iv)} form pathologies on the ECG-QA dataset. 
Success on these tasks requires models to integrate both high-level clinical reasoning about diagnostic patterns and low-level analysis of waveform morphology and signal characteristics.
For each of these tasks, the models are prompted to provide predictions along with the associated reasoning traces.
Subsequently, we pass the inputs and the reasoning traces through our \projectname pipeline to obtain the Perception and Deduction results.
We tabulate full results in the Appendix (\Cref{tab:comprehensive_results}) and show key metrics in \Cref{fig:overall_performance}.

\textbf{Results.} 
In \Cref{fig:overall_performance} (left), we plot Perception performance across tasks. 
For ECG-QA rhythm- and form-related symptoms, TSLMs (QoQ-Med and OpenTSLM) achieve the best \edit{peak} results, verifying 25--30\% of reasoning traces. 
\edit{Meanwhile, Gemini 3.1 Pro Plot demonstrates the most robust consistency, maintaining roughly 15--16\% Acc@Thresh100\% across all four datasets.}
However, Acc@Thresh100\% remains below 10\% for \edit{most} other cases, highlighting the difficulty of Perception. 
Relaxing the success threshold to verify at least 50\% of findings (Acc@Thresh50\%) substantially boosts performance to 70--80\% (\Cref{tab:comprehensive_results}). 
We approximate the Physician Performance upper bound using assessment data from Section \ref{sec:perception} rather than conducting infeasible large-scale manual annotation. 
No model approaches the perception level of a physician.

\Cref{fig:overall_performance} (right) shows that TSLMs underperform on ECG-QA tasks, indicating reasoning that aligns less with medical consensus. 
In contrast, the frontier model Claude Opus 4.5 produces more accurate deduction performance consistent across tasks. 
Notably, passing plotted ECG signals to Claude Opus 4.5 yields better results than using raw numbers, likely because general-purpose LLMs lack exposure to raw arrays.
\edit{Building on the success of vision-based inputs, Gemini 3.1 Pro Plot establishes a new state-of-the-art for deduction among the evaluated models. It substantially outperforms Claude Opus 4.5 Plot across all datasets, most notably reaching a Precision@5 of 0.48 on the Rhythm dataset.}
We again approximate the physician upper bound by averaging the performance of physician-defined criteria per dataset from Section \ref{sec:deduction}. 
Consistent with Perception results, physician-based annotations significantly outperform model-based reasoning, demonstrating a substantial gap in deductive capability.

\section{Discussion} \label{sec:discussion}

\textbf{Perception Flags Incorrect and Challenging Cardiologist Reports.}
Our Perception pipeline serves not only as a verification tool for model outputs but also as a quality assurance mechanism for human-annotated ground truth. In our analysis of Perception failures (Figure 3), we found that in 17\% of cases, the ``Code is right, Cardiologist note is wrong". 
This occurs when the cardiologist's report contains descriptors that are empirically absent from the signal, suggesting the ground truth itself may be flawed or imprecise. 
Various examples of this failure case with annotations can be found in Appendix (Figures \ref{fig:mimicwrong67}, \ref{fig:mimicwrong81}, \ref{fig:mimicwrong82}, \ref{fig:mimicwrong85}).

Furthermore, the category ``Code is maybe wrong" highlights the inherent difficulty of verifying reports that rely on external context; for instance, a description of ``within normal limits" cannot be verified without demographic data (e.g. patient age). 
The ability of our framework to categorize these discrepancies implies that our agentic perception mechanism is reliable enough to flag potential errors in expert clinical documentation.

\begin{figure}[!t]
    \vspace{-3mm}
    \centering
    \includegraphics[width=\linewidth]{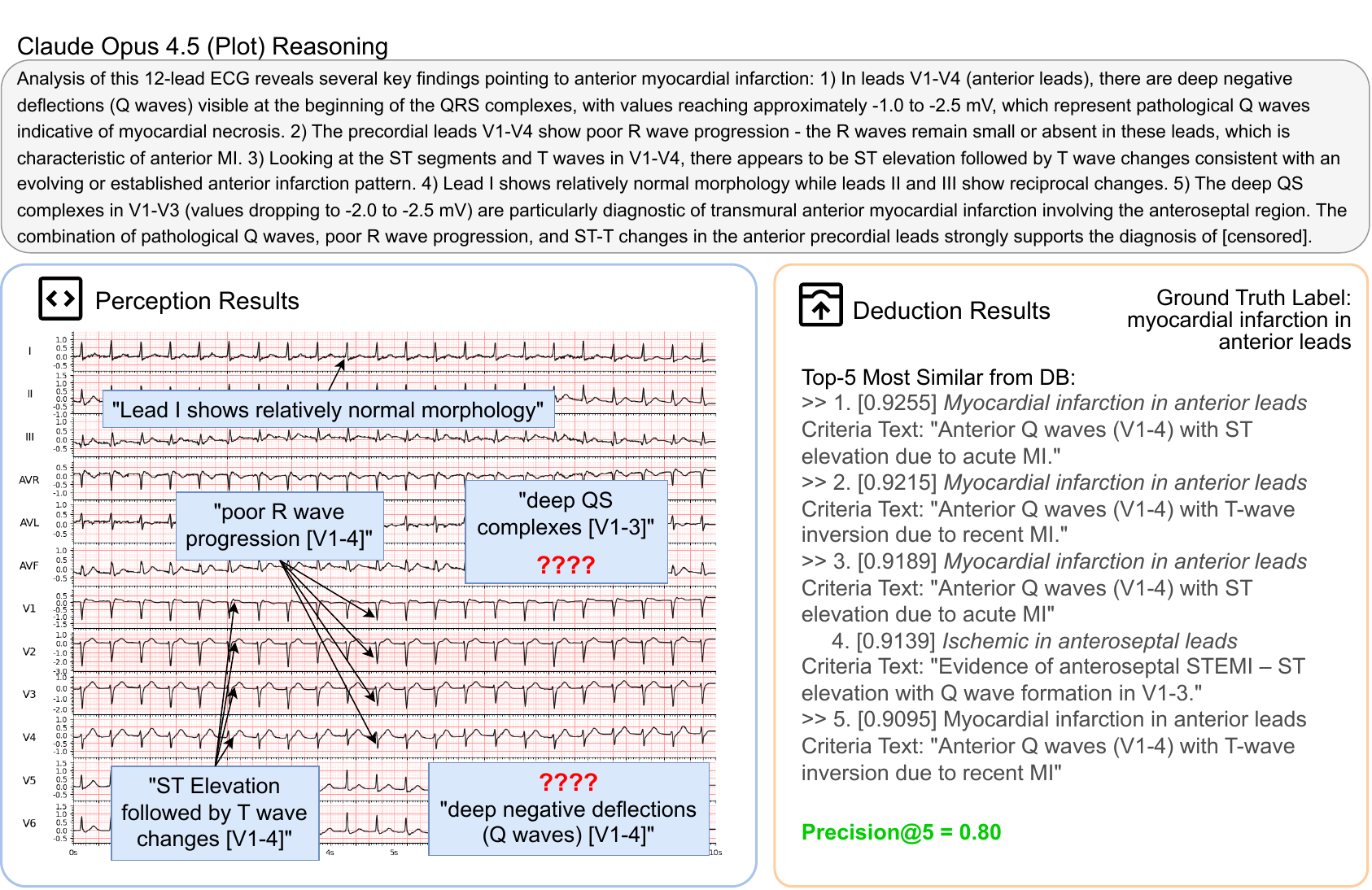}
    \vspace{-5mm}
    \caption{\textbf{Sample output of Claude (Plot).} Claude often produces correct diagnoses (high Deduction score), but with incorrect time-series reasoning (low Perception score). In this example, "deep QS complexes" are indicative of myocardial infarction, but do not appear in the ECG. This implies a post-hoc reasoning, in which the model first predicts the label then hallucinates justification. This is further implied by the structure of reasoning text, which first states the diagnosis then causally generates reasoning text.    
    \looseness -1
}
    \vspace{-5mm}
    \label{fig:model}
\end{figure}

\textbf{Deduction Works Well for a Range of Different Disease Presentations.}
Our Deduction evaluation demonstrates remarkable robustness to the stylistic variability of clinical reasoning. 
As shown in Figure \ref{fig:deductexamplephyscian}, physicians often arrive at the same diagnosis, but with different reasoning criteria and with vastly different linguistic structures. 
For the diagnosis of Incomplete Right Bundle Branch Block (IRBBB), Physician A provided a detailed, logic-heavy derivation by stating ``AND" and ``OR". 
In contrast, Physician B provided a terse, exclusionary description (``morphology not meeting criteria for complete RBBB") without explicitly defining the positive features of IRBBB. 
Despite Physician B’s description being very short and without much context, our retrieval pipeline successfully retrieved many top-k reference articles for IRBBB (see \Cref{fig:deduction_pipeline}). 
This confirms that our retrieval-based deduction metric focuses on the semantic clinical logic rather than surface-level keyword matching, making it effective even for atypical descriptions.

\textbf{Low Perception and High Deduction Makes an Untrustworthy Model.} 
A model that excels at Deduction but fails at Perception seems to exhibit a post-hoc reasoning mechanism, where it first predicts the correct label and then hallucinates a justification to fit that label. 
This phenomenon is most evident in Claude Opus 4.5 (Plot), which achieves strong performance on Deduction tasks and high global accuracy, yet suffers from poor Perception scores. This discrepancy suggests the model is fabricating diagnostic criteria to support its answer. \looseness -1

For example, in Figure 6, Claude correctly diagnoses Myocardial Infarction but justifies it by citing ``deep QS complexes" and ``deep Q waves" that are empirically non-existent in the signal. While these conditions are indeed clinically indicative of infarction, they cannot be valid reasoning if they are absent from the waveform. This creates a dangerous "illusion of thinking," where the model produces persuasive, textually accurate medical prose that is ungrounded in patient reality, posing serious trustworthiness concerns for clinical deployment. \looseness -1

\href{https://en.wikipedia.org/wiki/All_work_and_no_play_makes_Jack_a_dull_boy}{\textbf{\color{black}All Perception and No Deduction Makes the Model a Dull Boy.}}
Conversely, models with high Perception but low Deduction act as accurate sensors that lack the world knowledge to be clinically useful. This phenotype is observed in Time-Series Language Models (TSLMs) like OpenTSLM. \edit{
As shown in Figure 7, OpenTSLM successfully identifies the presence of inverted t-waves (Perception), showing that it can ``see" the signal features. 
However, it fails to identify that they are pathological and link it to the correct diagnosis (Left Ventricular Hypertrophy),} instead retrieving irrelevant clinical associations. While these models are grounded, they lack the broader medical domain knowledge required to differentially diagnose, limiting their utility to identifying features without understanding their implications. \looseness -1

\begin{figure}[!t]
    \vspace{-2mm}
    \centering
    \includegraphics[width=\linewidth]{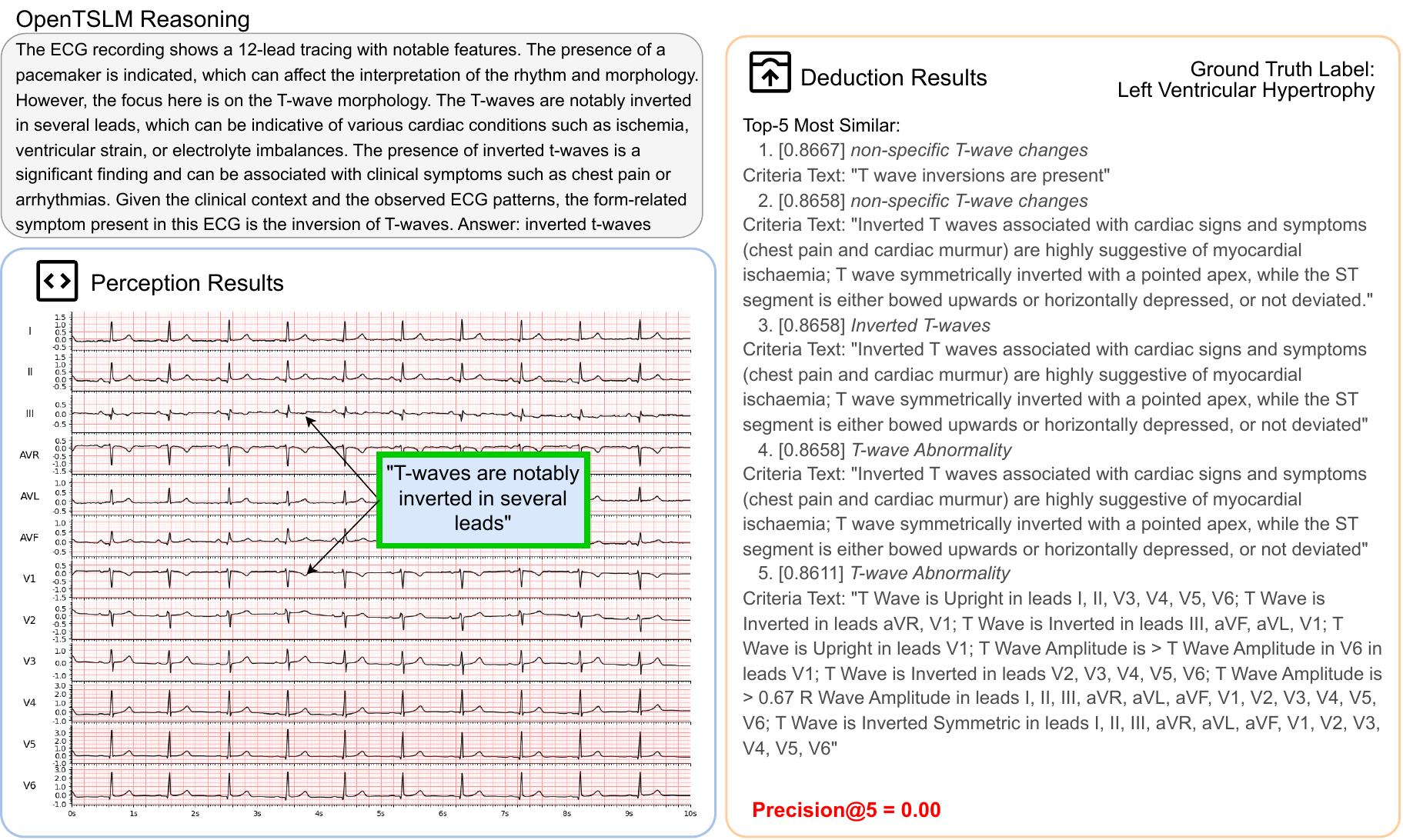}
    \vspace{-5mm}
    \caption{\textbf{Sample output of OpenTSLM.} While OpenTSLM generally has good Perception, their low Deduction performance implies that it suffers from being unable to identify the most salient features for precise diagnosis. In this example, although OpenTSLM correctly perceived the inverted t-waves, it failed to realize that these pathological features, in conjunction with the higher QRS voltage could indicate left ventricular hypertrophy, a disease with significant health implications. \looseness -1
    }
    \vspace{-5mm}
    \label{fig:model}
\end{figure}

\edit{\textbf{Frontier Multimodal Models Show Promise in Balancing Perception and Deduction.}
While TSLMs and Claude Opus 4.5 demonstrate isolated strengths in either perception or deduction, Gemini 3.1 Pro Plot begins to bridge this gap. As observed in our evaluations, Gemini 3.1 Pro Plot achieves the highest Deduction scores across all datasets (e.g., reaching a Precision@5 of 0.48 on the Rhythm dataset) while simultaneously maintaining the most consistent and highest Perception scores (approximately 15--16\% Acc@Thresh100\% across all tasks) among the evaluated models. This balanced performance suggests that advanced multimodal architectures can be better equipped to both ``see" the raw physiological signal and ``understand" its clinical implications without immediately resorting to post-hoc hallucination. However, because its perception capabilities still fall far short of the physician upper bound, the model cannot yet be treated as a fully trustworthy autonomous diagnostic agent.} \looseness -1

\textbf{Final Performance Accuracy Does Not Imply Reasoning Capability.}
Our results challenge the assumption that predictive accuracy serves as a proxy for reasoning quality. 
In \Cref{fig:correlation_plot}, we observe a strong correlation between Deduction and final accuracy (r=0.70), but a very weak correlation between Perception and final accuracy (r=0.18). 
This implies that models can achieve high classification performance without actually `seeing' the signal correctly. 
Furthermore, manual inspection reveals that when a model's final prediction is wrong, its Deduction is almost always wrong as well. 
This tight coupling suggests that high-performing models may simply be memorizing text-diagnosis pairs rather than engaging in true dual-process reasoning. 
Consequently, optimizing solely for final classification accuracy is suboptimal for developing models that can both describe a signal accurately and reason about it. 
Thus, while fully supervised models remain the optimal choice for maximizing predictive performance, the development of systems with genuine reasoning capabilities requires a distinct evaluation paradigm that does not conflate prediction success with reasoning validity. \looseness -1




\section{Conclusion} In this work, we introduced \projectname, a framework that decomposes reasoning into verifiable \textit{Perception} and \textit{Deduction} components. By operationalizing these axes through agentic verification and retrieval-based consensus, we established a reproducible standard that obviates the need for unscalable manual expert review. While demonstrated on ECGs, our formulation is fundamentally modality-agnostic. Future work should extend this paradigm to other health domains paving the way for trustworthy systems.

\newpage
\section*{Impact Statement}

This work introduces a scalable verification framework for multimodal healthcare AI, with a focus on the interpretability of ECG signals.
Our work addresses both societal and environmental challenges.

By decomposing reasoning into Deduction and Perception, our approach mitigates the risk of hallucinations, which are severely detrimental for medical applications. 
As such, it provides a pathway towards auditable AI systems in high stakes applications, where task performance is not the sole consideration but user trust is a key factor.

The current gold standard for verifying the reasoning logic of healthcare AI models involves manual review by medical professionals.
This process is not scalable and subject to the availability of the healthcare workforce.
Our agentic verification framework offers a scalable alternative, thereby reducing the load on this workforce while enabling rigorous auditing of model outputs.

The environmental costs are a growing concern in the AI and ML communities.
We note that deploying large models carries significant carbon footprint and our retrieval-based approach utilizes existing models rather than requiring continuous fine-tuning, and is a more compute and energy efficient alternative.

\bibliography{bibliography}

\bibliographystyle{icml2026}

\newpage
\appendix
\onecolumn

\newpage
\section{Experiments} \label{appendix:experiments}
\begin{table*}[!h]
\centering
\caption{\textbf{Comprehensive Evaluation on MIMIC-IV and ECG-QA.} We compare models across four datasets. Metrics are grouped by \textit{Perception} (labeling accuracy), \textit{Deduction} (retrieval quality), and \textit{Original} (downstream accuracy). All models are now evaluated across all available tasks.}
\label{tab:comprehensive_results}
\resizebox{\linewidth}{!}{%
\begin{tabular}{lcccccccccccccccccccccccccccc}
\toprule
\textbf{Model} & 
\multicolumn{7}{c}{\textbf{MIMIC-IV}} & 
\multicolumn{7}{c}{\textbf{ECG-QA Diagnosis}} & 
\multicolumn{7}{c}{\textbf{ECG-QA Rhythm}} & 
\multicolumn{7}{c}{\textbf{ECG-QA Form}} \\
\cmidrule(lr){2-8} \cmidrule(lr){9-15} \cmidrule(lr){16-22} \cmidrule(lr){23-29}

& \multicolumn{3}{c}{Perception} & \multicolumn{3}{c}{Deduction} & {Orig}
& \multicolumn{3}{c}{Perception} & \multicolumn{3}{c}{Deduction} & {Orig}
& \multicolumn{3}{c}{Perception} & \multicolumn{3}{c}{Deduction} & {Orig}
& \multicolumn{3}{c}{Perception} & \multicolumn{3}{c}{Deduction} & {Orig} \\
\cmidrule(lr){2-4} \cmidrule(lr){5-7} \cmidrule(lr){8-8}
\cmidrule(lr){9-11} \cmidrule(lr){12-14} \cmidrule(lr){15-15}
\cmidrule(lr){16-18} \cmidrule(lr){19-21} \cmidrule(lr){22-22}
\cmidrule(lr){23-25} \cmidrule(lr){26-28} \cmidrule(lr){29-29}

& \rotatebox{90}{Mac@50} & \rotatebox{90}{Mac@100} & \rotatebox{90}{Micro} 
& \rotatebox{90}{P@1} & \rotatebox{90}{P@5} & \rotatebox{90}{P@10} & \rotatebox{90}{Acc}
& \rotatebox{90}{Mac@50} & \rotatebox{90}{Mac@100} & \rotatebox{90}{Micro} 
& \rotatebox{90}{P@1} & \rotatebox{90}{P@5} & \rotatebox{90}{P@10} & \rotatebox{90}{Acc}
& \rotatebox{90}{Mac@50} & \rotatebox{90}{Mac@100} & \rotatebox{90}{Micro} 
& \rotatebox{90}{P@1} & \rotatebox{90}{P@5} & \rotatebox{90}{P@10} & \rotatebox{90}{Acc}
& \rotatebox{90}{Mac@50} & \rotatebox{90}{Mac@100} & \rotatebox{90}{Micro} 
& \rotatebox{90}{P@1} & \rotatebox{90}{P@5} & \rotatebox{90}{P@10} & \rotatebox{90}{Acc} \\
\midrule

OpenTSLM & 
0.63 & 0.07 & 0.52 & 0.12 & 0.12 & 0.13 & 0.01 & 
0.79 & 0.07 & 0.62 & 0.04 & 0.07 & 0.07 & 0.10 & 
0.68 & 0.09 & 0.56 & 0.20 & 0.21 & 0.20 & 0.15 & 
0.74 & 0.30 & 0.60 & 0.14 & 0.12 & 0.12 & 0.12 \\

QoQ-Med & 
0.55 & 0.03 & 0.47 & 0.11 & 0.12 & 0.12 & 0.14 & 
0.50 & 0.00 & 0.46 & 0.09 & 0.08 & 0.09 & 0.08 & 
0.79 & 0.25 & 0.69 & 0.21 & 0.19 & 0.18 & 0.14 & 
0.75 & 0.03 & 0.59 & 0.12 & 0.15 & 0.15 & 0.13 \\

Claude Array & 
0.83 & 0.09 & 0.66 & 0.15 & 0.15 & 0.16 & 0.16 & 
0.61 & 0.02 & 0.55 & 0.17 & 0.17 & 0.18 & 0.18 & 
0.73 & 0.02 & 0.61 & 0.22 & 0.22 & 0.20 & 0.15 & 
0.59 & 0.04 & 0.53 & 0.05 & 0.09 & 0.12 & 0.14 \\

Claude Plot & 
0.61 & 0.04 & 0.56 & 0.22 & 0.21 & 0.21 & 0.22 & 
0.71 & 0.05 & 0.59 & 0.22 & 0.20 & 0.20 & 0.22 & 
0.66 & 0.03 & 0.57 & 0.25 & 0.25 & 0.24 & 0.20 & 
0.56 & 0.06 & 0.52 & 0.18 & 0.21 & 0.22 & 0.21 \\

Gemini Plot & 
0.74 & 0.15 & 0.63 & 0.32 & 0.31 & 0.30 & 0.33 & 
0.77 & 0.16 & 0.67 & 0.28 & 0.27 & 0.27 & 0.28 & 
0.80 & 0.16 & 0.66 & 0.47 & 0.48 & 0.45 & 0.38 & 
0.74 & 0.16 & 0.64 & 0.28 & 0.32 & 0.32 & 0.39 \\

\bottomrule
\end{tabular}
}
\end{table*}

\begin{figure}[!htbp]
    \centering
    \includegraphics[width=\linewidth]{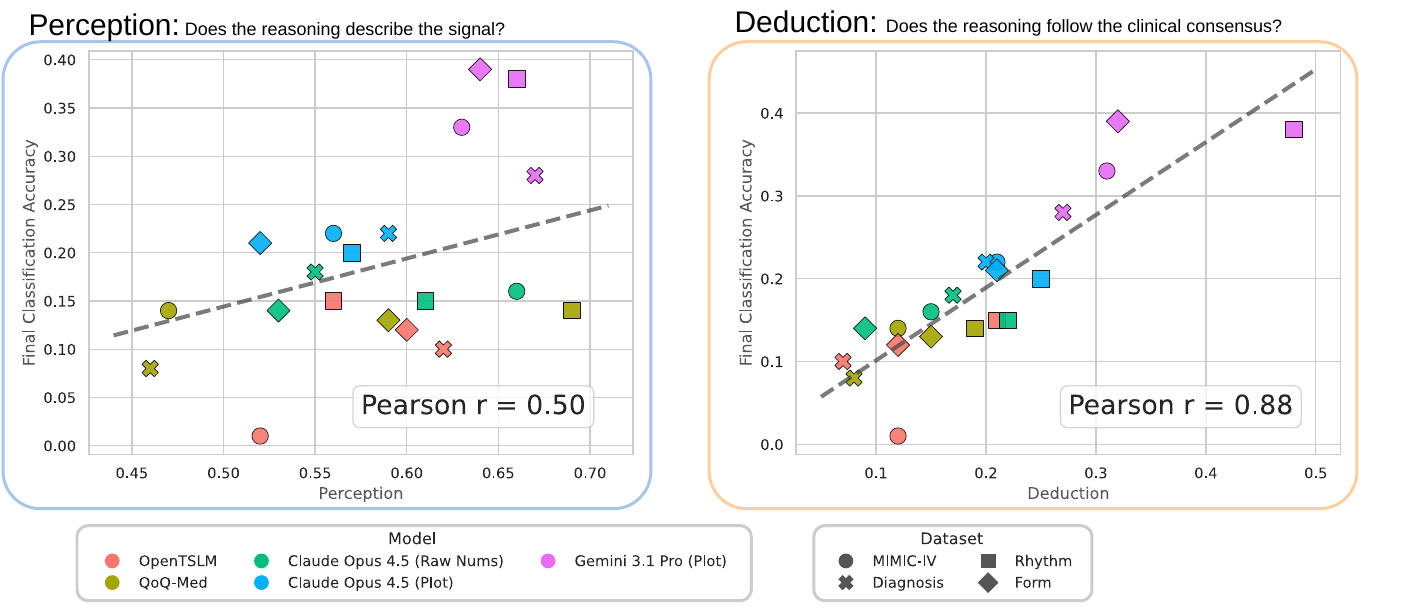}
    \caption{\textbf{Correlation with Prediction Accuracy.} Prior approaches have focused on benchmarking only on Final Classification Accuracy, without explicitly benchmarking the reasoning capability. Deduction is highly correlated to final accuracy, but Perception is not. This discrepancy indicates that high final accuracy does not imply that the model is reasoning correctly on the specific signal features that are present, leading to issues in trustworthiness.    
    }
    \label{fig:correlation_plot}
\end{figure}
\clearpage

\section{Adversarial Assessment}
\begin{table}[h]
    \centering
    \caption{\textbf{Antonym Mappings used during Adversarial Assessment.}}
    \label{tab:app:adv_map}
    \begin{tabular}{p{.65\linewidth} p{0.35\linewidth}}
    \toprule
    \textbf{Antonyms} & \textbf{Notes} \\
    \midrule
    Wide $\leftrightarrow$ Narrow & e.g. Wide QRS  \\ 
    Elevation $\leftrightarrow$ Depression & e.g. ST Elevation \\ 
    Regular $\leftrightarrow$ Irregular & e.g. Regular rhythm \\ 
    Upright $\leftrightarrow$ Inverted & e.g. Inverted P wave \\ 
    Prolonged $\leftrightarrow$ Shortened & e.g. Prolonged PR \\ 
    Left $\leftrightarrow$ Right & e.g. Left axis deviation \\ 
    High $\leftrightarrow$ Low & e.g. High QRS voltage \\ 
    Poor $\leftrightarrow$ Normal & e.g. Poor progression \\ 
    \{Absent, Flat\} $\leftrightarrow$ \{Present, Peak\} & e.g. P waves absent \\ 
    \{Fusion, Fused\} $\leftrightarrow$ \{Distinct, Capture\} & e.g. Fused T waves \\ 
    Sinus Rhythm $\leftrightarrow$ Sinus Bradycardia $\leftrightarrow$ Sinus Tachycardia $\leftrightarrow$ Atrial Fibrillation $\leftrightarrow$ Atrial Flutter $\leftrightarrow$ Junctional Rhythm & Different heart rate speeds and rhythms. \\

    \bottomrule
    \end{tabular}
\end{table}

\section{Knowledge Sources}
\label{sec:appendix:knowledge_sources}

The aggregation, parsing, and subsequent release of our structured database are strictly permissible under the open-access frameworks governing these sources. As detailed be;pw, LITFL , Wikipedia , and WikiEM operate under CC BY-NC-SA 4.0 or CC BY-SA 4.0, while ECGpedia operates under CC BY-NC 3.0 NL. These Creative Commons licenses explicitly grant permission to "adapt" (i.e., parse and restructure) and "share" the underlying content. By releasing our derived database as an open-source asset for non-commercial research with proper attribution, we fully comply with the \textit{ShareAlike} and \textit{NonCommercial} provisions, ensuring our work respects the legal terms and open-source ethos of the original contributors. Please see copies of the license in our codebase under the folder license/.

\begin{itemize}[leftmargin=*,noitemsep, topsep=0pt]
\item \href{https://litfl.com/ecg-library/diagnosis/}{{\color{black} LITFL ECG Library:}} Gold standard for visual ECG recognition and diagnosis, commonly used by cardiologists and electrophysiologists. Licensed under CC BY-NC-SA 4.0, allowing for free adaption and sharing.

\item \href{https://en.wikipedia.org}{{\color{black}Wikipedia:}} Represents the broadest tier of medical consensus with general definitions and widely accepted standard-of-care descriptions. Licensed under CC BY-SA 4.0, allowing for free adaption and sharing. 

\item \href{https://en.ecgpedia.org/}{{\color{black}ECGpedia:}} A specialized knowledge base dedicated designed for teaching with textbook and course materials. Licensed under CC BY-NC 3.0 NL, allowing for free adaption and sharing. 

\item \href{https://wikiem.org/wiki/Category:ECG}{{\color{black}WikiEM:}} Focused on Emergency Medicine, this resource captures rapid triage and decision-making protocols. Licensed under CC BY-SA 4.0, allowing for free adaption and sharing. 
\end{itemize}

\section{Task Details} \label{appendix:taskdetails}
We evaluate on subsets ECG-QA \citep{oh2023ecgqa} and MIMIC-IV-ECG \citep{gow2023mimicivecg}. We evaluate on subsets in order to avoid excessive computational expenditure, but we ensure that the subsets are representative and demonstrate the most challenging scenarios.

For ECG-QA, we choose to focus on Question Template IDs 8, 10, 11. These questions represent the most difficult questions in ECG-QA, as they are multiple choice with many choices. Most other ECG-QA questions used binary answers.

Question 8 was "What diagnostic symptoms does this ECG show, including uncertain symptoms?" with the below possible answers. 

['complete left bundle branch block', 'complete right bundle branch block', 'digitalis effect', 'first degree av block', 'incomplete left bundle branch block', 'incomplete right bundle branch block', 'ischemic in anterior leads', 'ischemic in anterolateral leads', 'ischemic in anteroseptal leads', 'ischemic in inferior leads', 'ischemic in inferolateral leads', 'ischemic in lateral leads', 'left anterior fascicular block', 'left atrial overload/enlargement', 'left posterior fascicular block', 'left ventricular hypertrophy', 'long qt-interval', 'myocardial infarction in anterior leads', 'myocardial infarction in anterolateral leads', 'myocardial infarction in anteroseptal leads', 'myocardial infarction in inferior leads', 'myocardial infarction in inferolateral leads', 'myocardial infarction in inferoposterior leads', 'myocardial infarction in inferoposterolateral leads', 'myocardial infarction in lateral leads', 'myocardial infarction in posterior leads', 'non-diagnostic t abnormalities', 'non-specific intraventricular conduction disturbance (block)', 'non-specific ischemic', 'non-specific st changes', 'none', 'right atrial overload/enlargement', 'subendocardial injury in anterolateral leads', 'subendocardial injury in anteroseptal leads', 'subendocardial injury in inferior leads', 'subendocardial injury in inferolateral leads', 'subendocardial injury in lateral leads']

Question 10 was "What rhythm-related symptoms does this ECG show?" with the below answers.

['atrial fibrillation', 'atrial flutter', 'bigeminal pattern (unknown origin, supraventricular, or ventricular)', 'none', 'normal functioning artificial pacemaker', 'sinus arrhythmia', 'sinus bradycardia', 'sinus rhythm', 'sinus tachycardia', 'supraventricular tachycardia']

Question 11 was "What form-related symptoms does this ECG show?" with the below answers.

['abnormal qrs', 'atrial premature complex', 'digitalis effect', 'high qrs voltage', 'inverted t-waves', 'long qt-interval', 'low amplitude t-wave', 'low qrs voltages in the frontal and horizontal leads', 'non-diagnostic t abnormalities', 'non-specific st changes', 'non-specific st depression', 'non-specific st elevation', 'non-specific t-wave changes', 'none', 'prolonged pr interval', 'q waves present', 't-wave abnormality', 'ventricular premature complex', 'voltage criteria (qrs) for left ventricular hypertrophy']

For MIMIC-IV-ECG, we parse out the ICD-10 diagnosis codes from the corresponding cardiologist notes and choose the classes with at least 100 ECGs associated with them, as well as having an overlap in classes with ECG-QA. This gives us these 8 classes for a total of 800 ECGs.

[left bundle branch block, supraventricular tachycardia, right bundle branch block, ventricular premature depolarization (pvcs), long qt syndrome, st segment elevation (stemi) myocardial infarction, atrial flutter, atrial fibrillation]

\begin{table}[h]
\centering
\caption{\edit{ICD-10 Codes and Corresponding Diagnoses used in MIMIC-IV evaluation.}}
\label{tab:icd10_codes}
\resizebox{0.95\linewidth}{!}{%
\begin{tabular}{ll}
\toprule
\textbf{ICD-10 Code} & \textbf{Diagnosis} \\
\midrule
I21.3; I21.3 & ST segment elevation (STEMI) myocardial infarction; Acute ST elevation myocardial infarction \\
I48.91 & Atrial fibrillation \\
I48.92; I48.4 & Atrial flutter; Typical atrial flutter \\
I45.81 & Long QT syndrome \\
I47.1 & Supraventricular tachycardia \\
I49.3 & Ventricular premature depolarization (PVCs) \\
I45.10 & Right bundle branch block \\
I44.7 & Left bundle branch block \\
\bottomrule
\end{tabular}%
}
\end{table}


\clearpage



\section{Incorrect MIMIC-IV-ECG Cardiologist Notes Flagged by \projectname Perception}
\label{sec:appendix:wrongmimic}

\begin{figure*}[h]
    \centering
    \includegraphics[width=\linewidth]{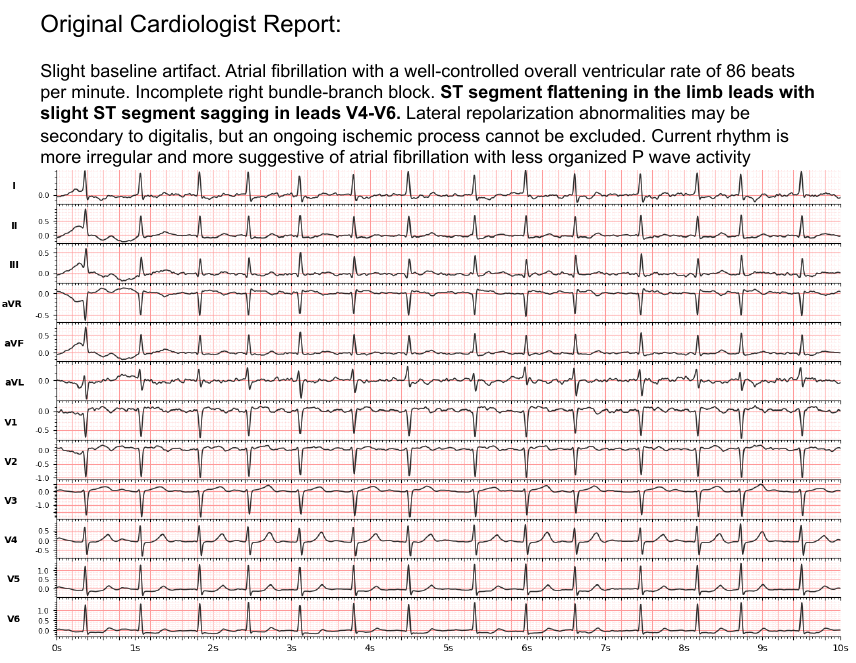}
    \vspace{-3mm}
    \caption{\textbf{MIMIC-IV cardiologist note flagged as being incorrect by \projectname, and verified to be incorrect by our cardiologists.} 
    Team's note: I don't agree with the ST changes in the limb leads, I think that's just artifact. I think it's ok to say non-specific ST changes in V5 and V6 but it's submillimeter in many areas so would also be fine to leave it out. The QRS morphology does not look like an incomplete RBBB even though the report says that. 
    }
    \label{fig:mimicwrong67}
\end{figure*}
\newpage

\begin{figure*}[h]
    \centering
    \includegraphics[width=\linewidth]{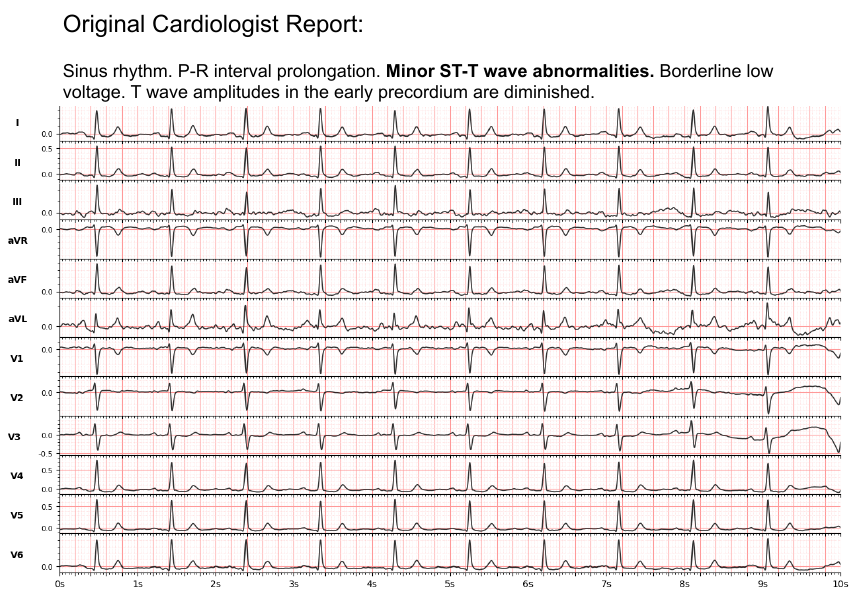}
    \vspace{-3mm}
    \caption{\textbf{MIMIC-IV cardiologist note flagged as being incorrect by \projectname, and verified to be incorrect by our cardiologists.}
    Team's note:  I wouldn't call minor ST abnormalities. It's just artifact, or maybe it thinks the TWI are abnormal in V1 and V2, but that can be a normal variant. 
    }
    \label{fig:mimicwrong81}
\end{figure*}
\newpage

\begin{figure*}[h]
    \centering
    \includegraphics[width=\linewidth]{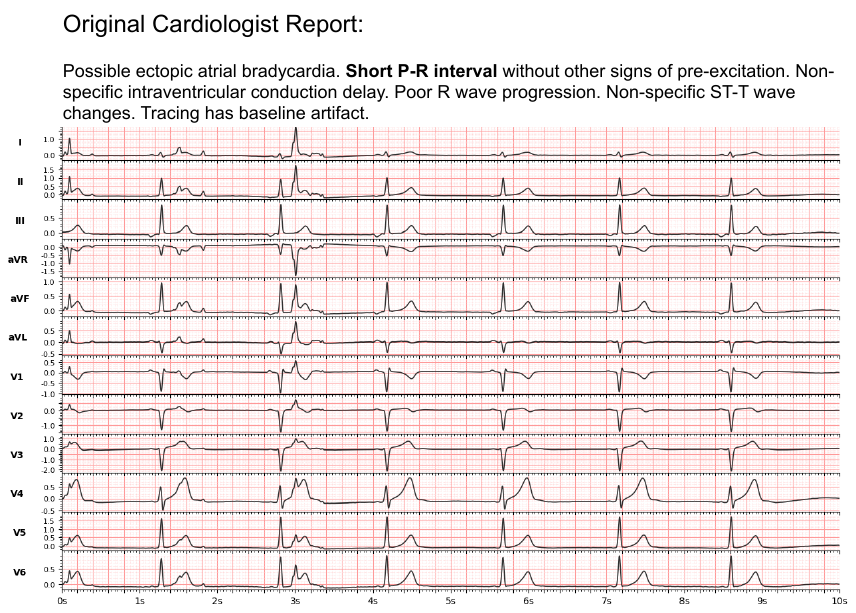}
    \vspace{-3mm}
    \caption{\textbf{MIMIC-IV cardiologist note flagged as being incorrect by \projectname, and verified to be incorrect by our cardiologists.} 
    Team's note: PR interval looks normal.
    }
    \label{fig:mimicwrong82}
\end{figure*}
\newpage

\begin{figure*}[h]
    \centering
    \includegraphics[width=\linewidth]{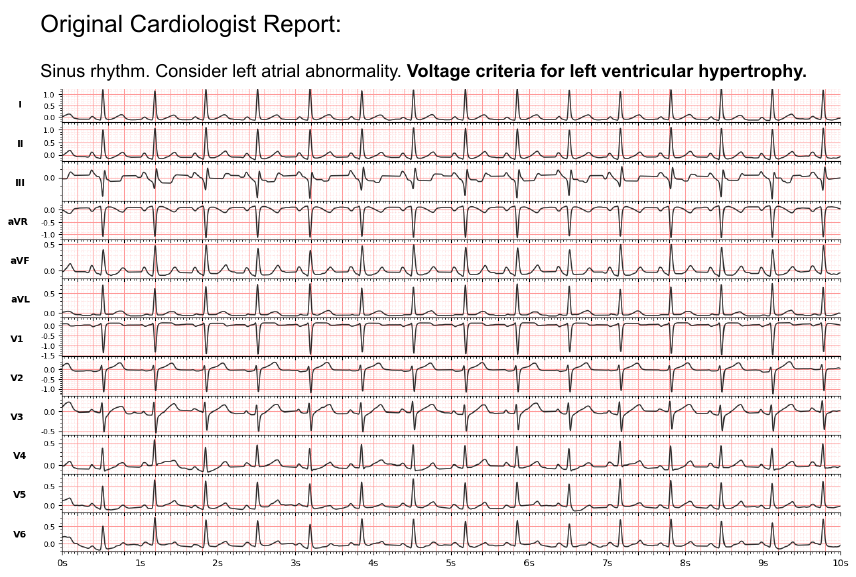}
    \vspace{-3mm}
    \caption{ 
    \textbf{MIMIC-IV cardiologist note flagged as being incorrect by \projectname, and verified to be incorrect by our cardiologists.} 
    Team's Note: Agreed, does not look like LVH, also don't see left axis deviation or strain pattern which would be expected with LVH.
    }
    \label{fig:mimicwrong85}
\end{figure*}
\newpage





\newpage

\clearpage
\section{Perception Phase Details}

\label{appendix:perception}

\begin{figure*}[h]
    \centering
    \includegraphics[width=0.7\linewidth]{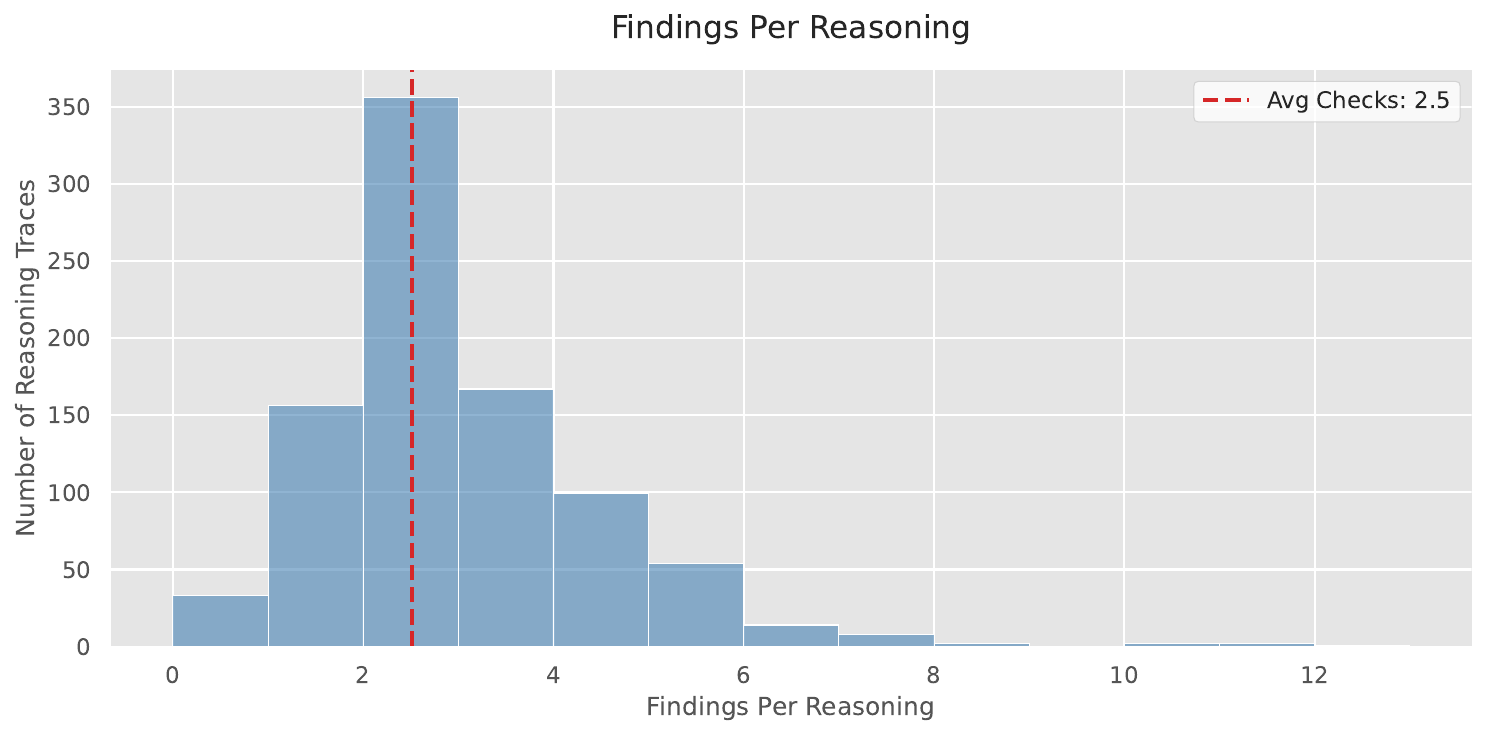}
    \caption{\textbf{Number of findings per trace.}
    }
    \label{fig:model}
\end{figure*}

\begin{figure*}[h]
    \centering
    \includegraphics[width=0.7\linewidth]{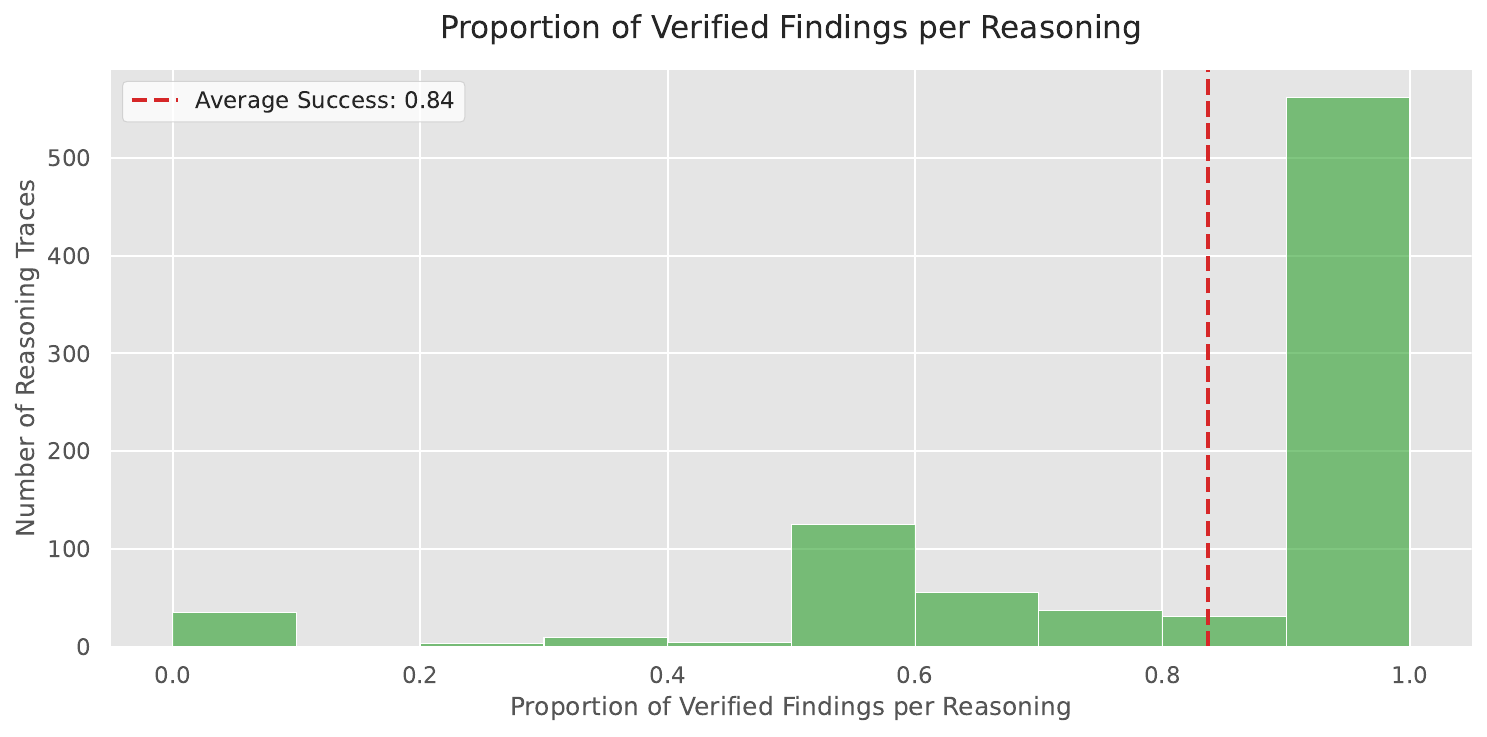}
    \caption{\textbf{Verified Findings per Trace.}
    }
    \label{fig:model}
\end{figure*}
\clearpage

\begin{figure*}[h]
    \centering
    \begin{tcolorbox}[
    title={\mbox{\scriptsize \normalfont Input: \texttt{reasoning\_trace}} \hfill \textbf{Perception Data Science Agent}},
    boxrule=0.8pt,
        arc=2mm,
        left=8pt, right=8pt, top=8pt, bottom=8pt
    ]
    \scriptsize
    \textbf{System Prompt:} You are an ECG analysis expert, Cardiac Electrophysiologist, and a data scientist.

    \medskip
    \textbf{Input Data:}
    \begin{itemize}[leftmargin=*, noitemsep, topsep=0pt]
        \item {ecg\_signal} (numpy.array) shape (12, 5000): 12-lead 10s ECG, fs=500Hz in this lead order \texttt{\{ecg\_leads\_available\}}.
        \item {ecg\_features\_dict} (Dict[str, Dict]): Pre-computed features extracted from an ECG segmentation algorithm.
        \begin{itemize}[leftmargin=*, noitemsep]
            \item The dictionary is keyed by lead name (e.g., 'I', 'V1'). Each value is a dictionary of features for that lead.
        \end{itemize}
    \end{itemize}

    \medskip
    \textbf{Data Schema and Structure:}
    \begin{itemize}[leftmargin=*, noitemsep, topsep=0pt]
        \item The inner dictionaries contain "Scalar Values" (floats) and "Array Values" (1D NumPy arrays).
        \item \textbf{Top-Level Keys Available (Leads):} ['I', 'II', 'III', 'aVR', 'aVF', 'aVL', 'V1', 'V2', 'V3', 'V4', 'V5', 'V6']
        \item \textbf{Inner Dictionary Keys (Features):}
        \begin{itemize}[leftmargin=*, noitemsep]
            \item {Scalar Values (float):} {avg\_PR\_interval\_(msec), avg\_QRS\_interval\_(msec), avg\_QT\_interval\_(msec), avg\_QTc\_interval\_(msec), avg\_RR\_interval\_(msec), avg\_heart\_rate\_(bpm), avg\_ST\_segment\_(msec), avg\_P\_peak\_amp\_(mv), avg\_QRS\_peak\_amp\_(mv), avg\_T\_peak\_amp\_(mv), avg\_ST\_deviation\_(mv)}
            \item {Array Values (np.ndarray[float]):} {RR\_intervals\_(msec)}
            \item {Array Values (np.ndarray[int]):} {P\_on\_idxs, P\_off\_idxs, QRS\_on\_idxs, QRS\_off\_idxs, T\_on\_idxs, T\_off\_idxs}
        \end{itemize}
    \end{itemize}
    \medskip
    \textbf{Example {ecg\_features\_dict} Construction:} \\
    Below is a runnable code snippet that creates a sample dictionary. It uses a {subset of the features listed above} but demonstrates the {exact data structure} you will receive.
     \vspace{-0.8em}
    \begin{minted}[fontsize=\tiny, baselinestretch=0.9, frame=single]{text}
import numpy as np
ecg_features_dict = {
    'I': {
        'avg_QRS_interval_(msec)': 71.0,  # Scalar float
        'P_on_idxs': np.array([198, 890, 1488, 2500]) # 1D NumPy array
    },
    'II': {
        'avg_QRS_interval_(msec)': 87.25,
        'P_on_idxs': np.array([197, 889, 1488, 2502])
    }
}
    \end{minted}

    \textbf{Target Reasoning Trace:} \texttt{\{reasoning\_trace\}}

    \medskip
    \textbf{Your Task:}
    \begin{enumerate}[leftmargin=*, noitemsep, topsep=0pt]
        \item \textbf{Extract Findings (JSON Format):} Parse the reasoning trace and extract every specific, verifiable ECG finding. If anything is non-specific, such as "Artifact is present", then ignore the finding. Ignore any "diagnostic" conditions, such as "Right bundle-branch block". Focus on rhythm or form properties of the ECG signal. Also, ignore any findings that have to do with a pacemaker, such as "paced rhythm". Output this as a {JSON Dictionary} inside a \texttt{json} code block. Each dictionary must have: \texttt{key} (str): finding name; \texttt{value} (List[str]): exact quotes from the trace.
        \item \textbf{Write Verification Function (Python Format):} Write a Python function \texttt{verify\_perception(ecg\_signal, ecg\_features\_dict)} inside a \texttt{python} code block. Return a dictionary \texttt{findings\_checks} where each key are the specific claims from the JSON dictionary and values are the boolean result.
        \end{enumerate}
    
    \medskip
     \textbf{Output JSON Format Example:} \vspace{-0.8em}
        \begin{minted}[fontsize=\tiny, baselinestretch=0.9, frame=single, style=bw]{text}
{"Sinus Tachycardia": ["Heart rate is mentioned as >100 bpm"],
 "ST Elevation in V1": ["there is an ST Elevation in V1..."]}
        \end{minted}

    \medskip
    \textbf{Output Python Format Function Example:}  \vspace{-0.8em}
        \begin{minted}[fontsize=\tiny, baselinestretch=0.9, frame=single, style=bw]{text}
def verify_perception(ecg_signal, ecg_features_dict):
    import numpy as np
    findings_checks = {}
    hr = ecg_features_dict['II']['avg_heart_rate_(bpm)'].mean()
    findings_checks['Sinus Tachycardia'] = bool(hr > 100)
    return findings_checks
        \end{minted}
    \end{tcolorbox}
    \caption{\textbf{Verbatim System Prompt Specification for the Perception Verification Agent.}}
    \label{fig:full_system_prompt}
\end{figure*}

\begin{figure*}[h]
    \centering
    \begin{tcolorbox}[
        title={\mbox{\scriptsize \normalfont Input: \texttt{error\_trace}} \hfill \textbf{Perception Data Science Agent, Debugging}},
        boxrule=0.8pt,
        arc=2mm,
        left=8pt, right=8pt, top=8pt, bottom=8pt
    ]
    \scriptsize
    \textbf{System Prompt:} The previous function execution failed. Please analyze the error trace and fix the \texttt{verify\_perception} function.

    \medskip
    \textbf{Your Task:}
    \begin{enumerate}[leftmargin=*, noitemsep, topsep=0pt]
        \item \textbf{Analyze} the error trace to determine why the code failed (e.g., shape mismatch, invalid key access, logic error).
        \item \textbf{Re-write} the full Python function \texttt{verify\_perception(ecg\_signal, ecg\_features\_dict)} to fix the issue. Explain what the issue was as well.
        \item \textbf{Abstractness Check:} If the reasoning trace is truly too abstract to verify with the given data (e.g., requires historical patient data not in the inputs), return ONLY the text \texttt{"Perception is too abstract: \{REASONING\}"} with REASONING filled out and do not generate code.
    \end{enumerate}

    \medskip
    \textbf{Considerations:}
    \begin{itemize}[leftmargin=*, noitemsep, topsep=0pt]
        \item Do not infer or add checks for other criteria that might be part of a full diagnostic algorithm unless they are explicitly stated in the reasoning trace. The function must be a direct and literal translation of the text's claims into code.
        \item Do not put any \texttt{try-except} blocks in the code. If the code is written incorrectly, an error should be thrown to make it clear that the code must be fixed.
        \item Do not add a \texttt{main} function; only return the \texttt{verify\_perception} function or the "too abstract" string.
    \end{itemize}

    \medskip
    \textbf{Error Trace:} \\
    \texttt{\{error\_trace\}}
    
    \end{tcolorbox}
    \caption{\textbf{System Prompt for the Debugging Agent.} This agent is triggered when the initial code execution fails, tasked with either repairing the Python function or flagging the claim as unverifiable.}
    \label{fig:debug_prompt}
\end{figure*}

\clearpage
\section{Deduction Phase Details} \label{appendix:deduction}
\begin{figure}[h]
    \centering
    \includegraphics[width=\linewidth]{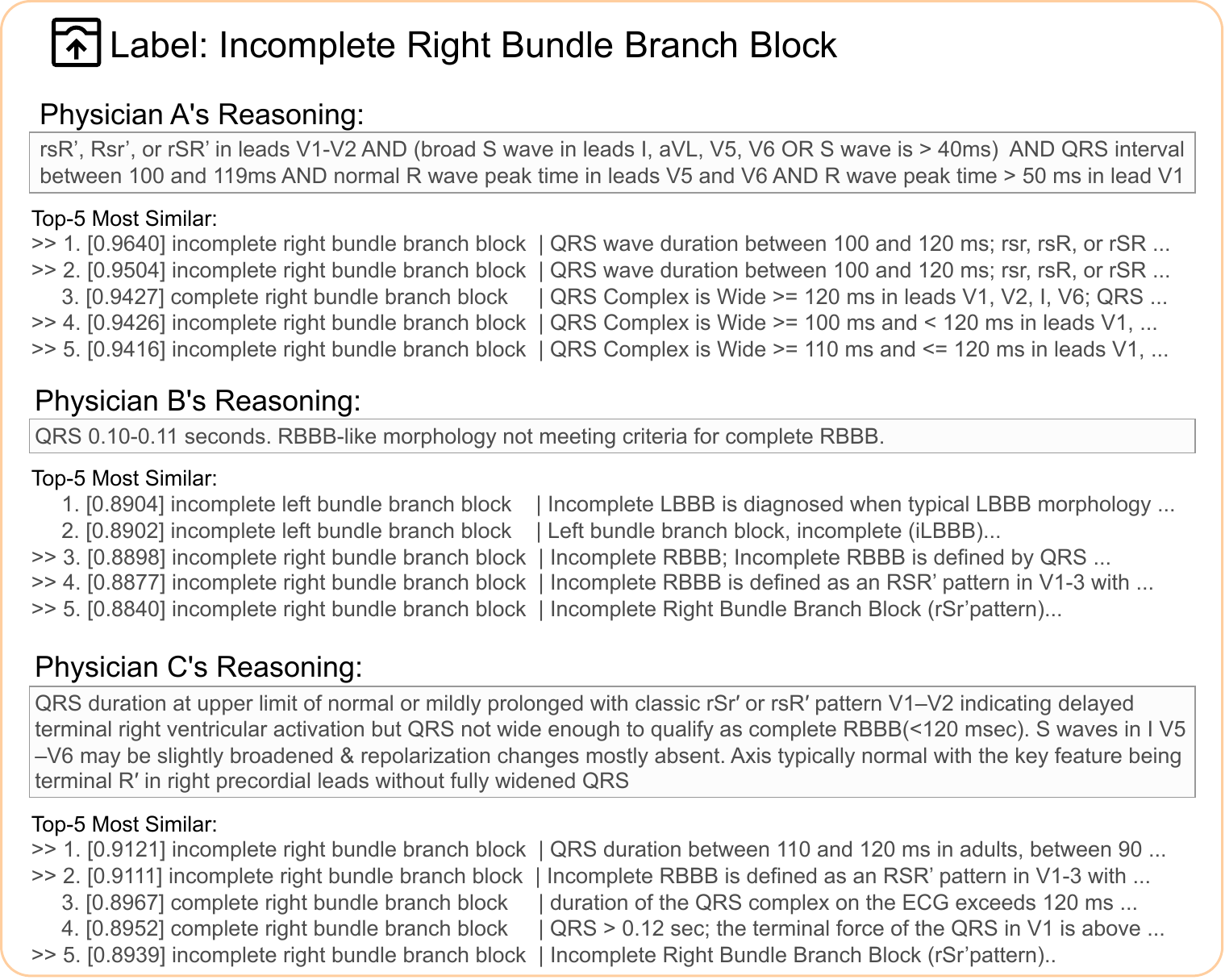}
    \vspace{-3mm}
    \caption{\textbf{Example of our Deduction Pipeline.} This example is how we validate our deduction, where we embed the Physician's reasoning, then compare it against our database to find the Top-$k$ most similar diagnostic criteria. Here, for Incomplete Right Bundle Block, we are able to consistently get at least 3 database entries on IRBBB based on the phsyician's definition, despite each definition being so differnetly framed. In particular, Physician B is very short, and doesnt even describe what RBBB is in it, just that IRBBB is not RBBB, but yet, we are still able to identify IRBBB articles. During inference, instead of a physician reasoning being used as input, the evaluated model's generated reasoning is used and compared to our database. 
    }
    \label{fig:deductexamplephyscian}
\end{figure}

\begin{figure*}[t!]
    \centering
    \begin{tcolorbox}[
    title={\mbox{\scriptsize \normalfont Input: \texttt{raw\_text}, \texttt{diagnosis}} \hfill \textbf{Text Cleaning Agent, Quotes}},
        boxrule=0.8pt,
        arc=2mm,
        left=8pt, right=8pt, top=8pt, bottom=8pt
    ]
    \scriptsize
    \textbf{System Prompt:} You are an Expert Medical Data Extractor. Your goal is to identify every distinct \textbf{Diagnostic Pattern, Class, or Criteria Group} from an ECG signal associated with {\{\texttt{diagnosis}\}} from the input text below. The input text is scraped from online resources on ECGs, so not all of the information may be useful or relevant to \{\texttt{diagnosis}\}.

    \medskip
    \textbf{CRITICAL INSTRUCTION: STRICT RELEVANCE}
    \begin{itemize}[leftmargin=*, noitemsep, topsep=0pt]
        \item \textbf{Strict Association:} Must be diagnostic criteria associated with {\{\texttt{diagnosis}\}}.
        \item \textbf{No Matches:} If the provided input text does NOT contain specific diagnostic criteria for {\{\texttt{diagnosis}\}}, you must return an empty list for "diagnostic\_clusters". 
        \item \textbf{Do Not Hallucinate:} Do not use outside medical knowledge to fill in gaps. Only extract what is present in the provided text.
        \item \textbf{Handling Types and Variants:} Extract EACH "Type" (e.g., Type 1) or "Pattern" (e.g., Coved vs. Saddleback) as a separate group, even if noted as "non-diagnostic".
    \end{itemize}

    \medskip
    \textbf{GROUPING LOGIC}
    \begin{enumerate}[leftmargin=*, noitemsep, topsep=0pt]
        \item \textbf{Explicit Types/Classes:} Each explicit label (e.g., "Classic Form") defines a separate group.
        \item \textbf{"OR" Logic:} If the text implies "Diagnosis is made by finding A... OR by finding B", these are separate groups.
        \item \textbf{Morphological Variants:} If the text describes alternative shapes (e.g., "instead of X, you see Y"), treat them as separate groups.
    \end{enumerate}

    \medskip
    \textbf{EXTRACTION RULES \& EXCLUSION CRITERIA}
    \begin{itemize}[leftmargin=*, noitemsep, topsep=0pt]
        \item \textbf{Verbatim Extraction:} Extract the \textbf{EXACT} phrases. Do not paraphrase.
        \item \textbf{Ignore Genetic/Etiological Lists:} Do NOT create groups for lists of mutations (e.g., SCN5A) UNLESS the text provides a \textit{distinct set of clinical criteria} specific to that gene.
        \item \textbf{Ignore non-specific criteria:} Ignore criteria not specific to \{diagnosis\}.
    \end{itemize}

    \medskip
    \textbf{INPUT TEXT:} \\
    \texttt{\{\texttt{raw\_text}\}}
    
    \medskip

    \textbf{OUTPUT FORMAT (Valid Raw JSON Only):} \vspace{-0.8em}
    \begin{minted}[fontsize=\tiny, baselinestretch=0.9, frame=single, style=bw]{text}
{
  "diagnostic_clusters": [
    {
      "concept_label": "Type 1 (Coved Pattern)", 
      "criteria": ["Coved ST segment elevation >2mm"]
    }
  ]
}
    \end{minted}
    \end{tcolorbox}
    \caption{\textbf{System Prompt Specification for the Medical Knowledge Extraction Agent.} This agent distills unstructured medical literature into structured diagnostic clusters for cross-verification against perception-level findings.}
    \label{fig:extraction_prompt}
\end{figure*}

\begin{figure*}[t!]
    \centering
    \begin{tcolorbox}[
        title={\mbox{\scriptsize \normalfont Input: \texttt{raw\_text}, \texttt{diagnosis}} \hfill \textbf{Text Cleaning Agent, Structured}},
        boxrule=0.8pt,
        arc=2mm,
        left=8pt, right=8pt, top=8pt, bottom=8pt,
        fonttitle=\small
    ]
    \scriptsize
    \textbf{System Prompt:} You are an Expert ECG Data Standardizer. Your goal is to extract diagnostic criteria for \textbf{\{diagnosis\}} from the input text and rewrite them using a \textbf{Strict Controlled Vocabulary Template}.


    \medskip
    \textbf{STANDARDIZATION RULES}
    \begin{enumerate}[leftmargin=*, noitemsep, topsep=2pt]
        \item \textbf{Strict Template:} Rewrite every criterion using this structure: \texttt{[Feature] is [Morphology] [Operator] [Value][Unit] in leads [Leads]}.
        \item \textbf{Standardize Logic:} Convert "greater than" to \texttt{>}, "at least" to \texttt{>=}. Convert time to \textbf{ms} and amplitude to \textbf{mm}.
        \item \textbf{Expand Leads:} Convert group names (e.g., "inferior leads") to specific leads (e.g., "II, III, aVF").
        \item \textbf{Relevance Check:} Extract criteria ONLY if they are specific to \textbf{\{diagnosis\}}.
    \end{enumerate}

    \medskip
    \textbf{INPUT TEXT:} \\
    \texttt{\{raw\_text\}}
    
    \medskip

    \textbf{OUTPUT FORMAT (Valid Raw JSON Only):} \vspace{-0.8em}
    \begin{minted}[fontsize=\tiny, baselinestretch=0.9, frame=single, style=bw]{text}
{
  "diagnostic_clusters": [
    {
      "concept_label": "Type 1", 
      "criteria": [
        "ST Segment is Elevated > 2mm in leads V1, V2"
      ]
    }
  ]
}
    \end{minted}
    \end{tcolorbox}
    \caption{\textbf{System Prompt Specification for the ECG Data Standardization Agent.} This agent enforces a controlled vocabulary and lead-expansion logic to ensure extraction results are machine-verifiable against numeric features.}
    \label{fig:standardization_prompt}
\end{figure*}

\end{document}